\documentclass[10pt,twocolumn,letterpaper]{article}
\usepackage[pagenumbers]{cvpr}
\usepackage{adjustbox}
\usepackage{tcolorbox}
\usepackage{stfloats}
\usepackage{algorithm}
\usepackage{algpseudocode}

\usepackage{amsmath,amsfonts,bm}

\def\eqref#1{equation~\ref{#1}}

\def\1{\bm{1}}

\DeclareMathAlphabet{\mathsfit}{\encodingdefault}{\sfdefault}{m}{sl}
\SetMathAlphabet{\mathsfit}{bold}{\encodingdefault}{\sfdefault}{bx}{n}

\usepackage[colorlinks=true,linkcolor=MidnightBlue,citecolor=MidnightBlue,urlcolor=MidnightBlue]{hyperref}
\newcommand{\lawtablefont}{\fontsize{9.73}{11.2}\selectfont}
\definecolor{lawabstractbg}{HTML}{F6F5FF}
\title{LAW: Unifying Language, Action, and World Modeling for Agentic Physical Intelligence}
\author{}

\let\lawincludegraphics\includegraphics
\renewcommand{\includegraphics}[2][]{\adjustbox{max width=\linewidth,max height=0.72\textheight}{\lawincludegraphics[#1]{#2}}}

\begin{document}
\twocolumn[{
\begin{minipage}{\textwidth}
\centering
\includegraphics[height=0.30in]{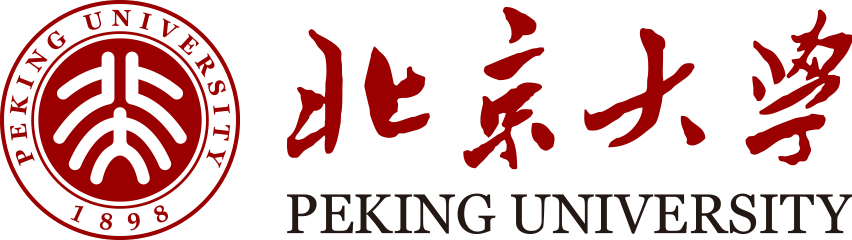}\hfill
\includegraphics[height=0.27in]{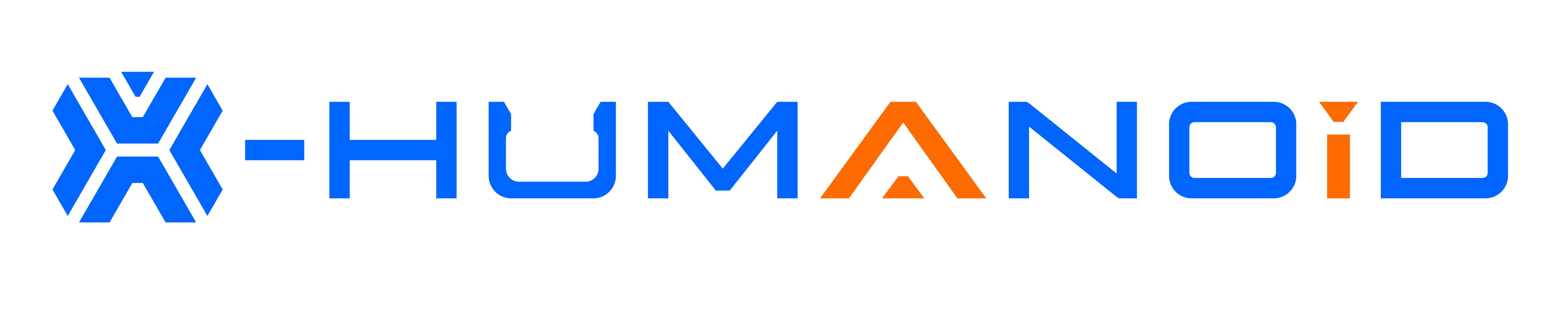}
\par\vspace{5pt}
\hrule height 0.6pt
\vspace{8pt}
\begin{tcolorbox}[colback=lawabstractbg,colframe=lawabstractbg,boxrule=0pt,arc=10pt,outer arc=10pt,left=12pt,right=12pt,top=12pt,bottom=12pt]
{\centering\LARGE\bfseries LAW: Unifying Language, Action, and World Modeling for Agentic Physical Intelligence\par}
\vspace{10pt}
{\raggedright\small
Jiajun Cao$^{1,2,*}$, Kai Tang$^{1,2,*}$, Bohua Zhang$^{1,2,*}$, Zezhong Qian$^{1,2,*}$, Ying Li$^{1,2,*}$, Xiaozhu Ju$^{2,*}$, Xiaobao Wei$^{1,2}$, Jiaming Liu$^{1}$, Yang Ding$^{1,2}$, Sixiang Chen$^{1,2}$, Yiding Sun$^{1,2}$, Siyuan Qian$^{1,2}$, Tinghao Wang$^{1,2}$, Jixian Wu$^{1,2}$, Yichen Guo$^{1,2}$, Yi Zhang$^{2}$, Shilong Zou$^{2}$, Zhiyuan Qin$^{2}$, Yong Dai$^{2}$, Jian Tang$^{2,\ddagger}$, Shanghang Zhang$^{1,\ddagger}$\par
\vspace{5pt}
$^{1}$State Key Laboratory of Multimedia Information Processing,
School of Computer Science, \\Peking University\\
$^{2}$Beijing Innovation Center of Humanoid Robotics\par
\vspace{3pt}
{\footnotesize $^{*}$Equal Contribution,\quad $^{\ddagger}$Equal Corresponding Author\par}}
\vspace{9pt}
{\normalsize\bfseries Abstract}\par\vspace{3pt}
\small World-action models jointly predict actions and future world states, but long-horizon manipulation further requires language reasoning to plan subtask objectives, assess execution progress, and guide action selection. We present LAW, a closed-loop VLM--WAM framework that unifies language reasoning, action generation, and world modeling through semantic subtask progress. Within the LAW agent, a hybrid-expert vision-language model plans the remaining subtasks, estimates the values of and fuses actions proposed by a world-action model, and asynchronously assesses execution status to support continued execution and subtask transitions. Within the world-action model, Mixture-of-Transformer Experts integrates modality-specific and shared computation for joint video--action modeling. Dual-stream hierarchical visual memory maintains temporal context for both models, while subtask-local value learning aligns action evaluation with semantic objectives and preserves supervision from successful prefixes of failed trajectories. LAW achieves 94.21\% average success across 50 RoboTwin tasks. Replacing the WAM backbone with OpenWAM-$\alpha$ further raises the average success rate to 95.56\%, establishing a new open-source state of the art on RoboTwin. Despite having only 2B parameters, our VLM also outperforms variants using Embodied-R1.5 and GPT-5.5 High.  On three long-horizon real-robot tasks, LAW achieves 65\% average success, compared with 28.33\% for $\pi_{0.5}$ and 25\% for Fast-WAM. \par\vspace{5pt}\noindent\textbf{Webpage:} \url{https://hey-cjj.github.io/LAW-project/}\\
\textbf{Code:} \url{https://github.com/hey-cjj/LAW}
\end{tcolorbox}
\end{minipage}
\vspace{12pt}
}]

\begin{figure*}[!tb]
\centering
\includegraphics[width=1.0\textwidth]{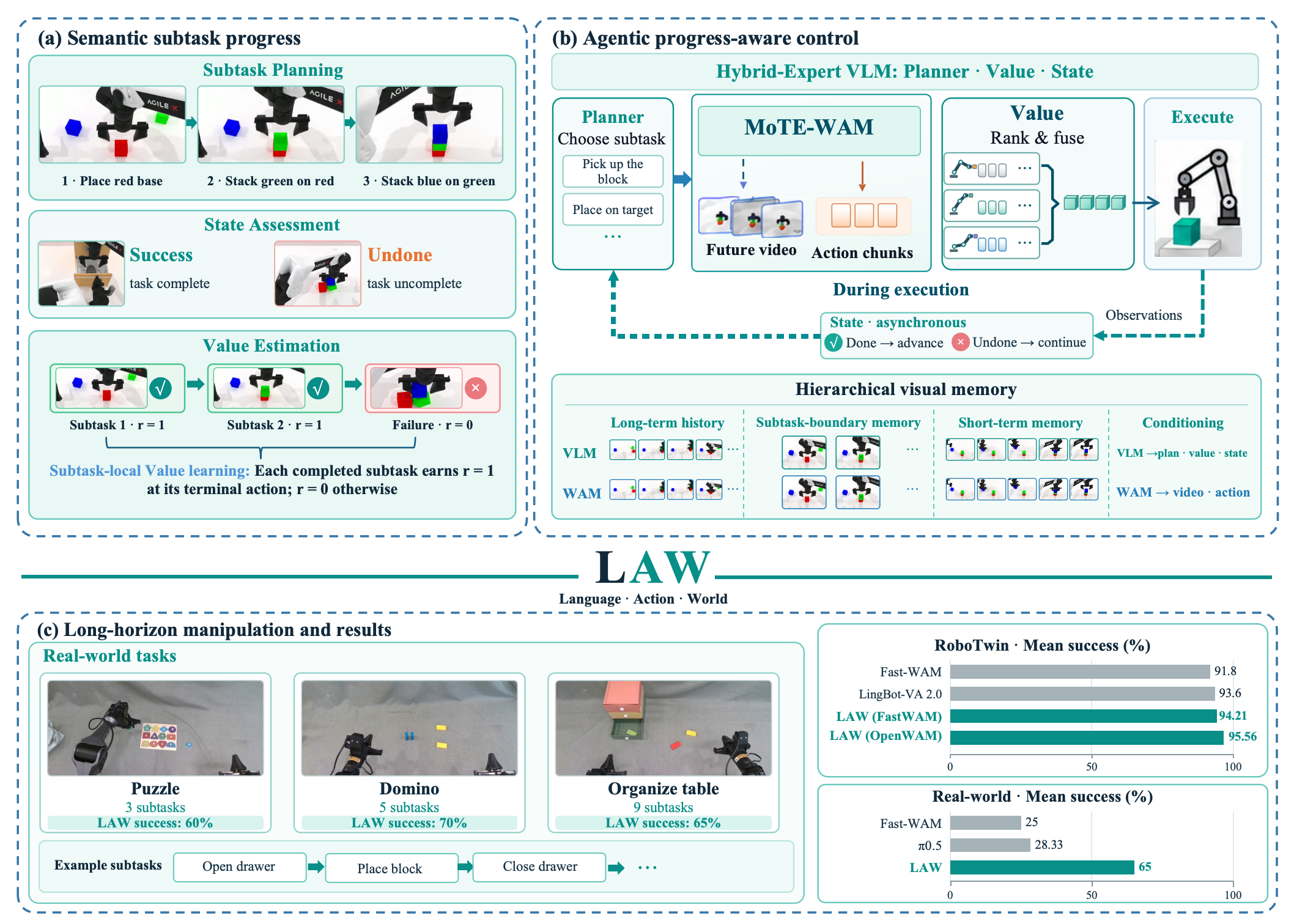}
\caption{Overview of the LAW agent.}
\label{fig:law-teaser}
\end{figure*}

\section{Introduction}
Long-horizon manipulation is a sequence of semantic subtasks, not merely a longer sequence of motor commands. After reaching for an object, a robot must decide whether the grasp actually succeeded; after a successful grasp, it must recognize that the task has entered a new stage; before executing the next motion, it must choose among several physically plausible actions according to the current objective~\cite{ahn2022saycan,huang2022innermonologue}. Errors in any of these decisions compound over time. Consequently, local action quality alone does not determine task success. A capable agent must maintain a persistent account of task progress, connect that account to action generation, and revise its decisions when execution departs from expectation~\cite{liu2023reflect,torne2026mem}.

Vision-language-action policies connect semantic representations to control~\cite{zitkovich2023rt2,kim2024openvla,black2024pizero}, while world-action models learn action-conditioned scene evolution and executable behavior~\cite{zhu2025uwm,zheng2025flare,zhang2026lingbotva,yuan2026fastwam}. These models substantially improve action generation, but long-horizon deployment exposes a complementary systems problem: reasoning and control must remain aligned as the task evolves. A language plan generated once at the beginning cannot determine whether its intermediate steps were completed~\cite{huang2022innermonologue}, and an action generator conditioned only on the current observation may not know which visual changes matter to the larger goal~\cite{shi2026memoryvla}. Placing a planner before a policy therefore leaves three interfaces underspecified: how execution progress updates the plan, how candidate actions are judged against the current semantic objective, and how relevant history is retained without overwhelming either model.

Our premise is that semantic subtask progress should serve as the temporal interface between language reasoning and world-action modeling. Subtasks provide more than a decomposition of the instruction: they define when the agent should advance its plan, which outcomes a candidate action should optimize, and which observations should remain salient in memory. This view reveals that computations occur at different rates. State assessment should run continuously without blocking control, candidate evaluation should occur whenever the WAM proposes a new action chunk, and planning should operate at the semantic task scale rather than at every control step. The challenge is coordinating reasoning and control, rather than invoking a larger reasoner at every step.

We instantiate this principle in LAW, a closed-loop VLM--WAM framework that unifies language reasoning, action generation, and world modeling through semantic subtask progress, as summarized in Fig.~\ref{fig:law-teaser}. The LAW agent uses a hybrid-expert VLM with role-selected planner, state, and value experts on a shared frozen backbone. The planner predicts the remaining semantic subtasks; conditioned on the current subtask, the WAM generates diverse action chunks; the value expert ranks and fuses them; and execution status is assessed asynchronously to determine subtask completion. Once the current subtask is completed, the planner predicts the next subtask and the agent switches to it; otherwise, the agent continues executing the current subtask. The LAW agent coordinates planning, action selection, and execution monitoring at different frequencies, using feedback to continue or switch subtasks. 

To strengthen joint world and action modeling, we introduce Mixture-of-Transformer Experts (MoTE) into the WAM, yielding MoTE-WAM. MoT projects video and action features into a common attention space to exchange cross-modal information, but their subsequent transformations remain modality-specific. MoTE introduces shared experts to further transform these cross-modal representations, alongside video-private and action-private experts that preserve modality-specific computation. The shared experts learn common temporal and physical structure across predicted scene dynamics and robot actions, strengthening video--action joint modeling beyond information exchange through attention alone. Long-horizon execution also requires memory aligned with subtask transitions. Separate hierarchical memory streams for the VLM and WAM retain recent observations and full-resolution states at subtask boundaries, while compressing older history to preserve relevant context across subtasks. To better exploit failed trajectories when training the value expert, we introduce subtask-local value learning, which terminates returns at semantic boundaries. This preserves positive supervision from successfully completed subtasks within failed episodes, rather than assigning zero value to all their actions. 

We evaluate LAW on 50 RoboTwin~\cite{chen2025robotwin} tasks, the RoboDojo~\cite{chen2026robodojo} capability benchmark, and three long-horizon real-robot tasks containing up to nine semantic subtasks. LAW achieves 94.21\% average success on RoboTwin, 2.38 percentage points above Fast-WAM, without robot-data pretraining. Replacing the WAM backbone with OpenWAM-$\alpha$ further raises the average success rate to 95.56\%, establishing a new open-source state of the art on RoboTwin. Despite having only 2B parameters, our VLM outperforms variants using Embodied-R1.5 and GPT-5.5 High within the same LAW framework. On RoboDojo, LAW achieves leading performance across capability dimensions among methods without additional robot-domain or embodied-interaction pretraining. On the three long-horizon real-robot tasks, LAW achieves 65\% average success, compared with 28.33\% for $\pi_{0.5}$ and 25\% for Fast-WAM. We also conduct extensive ablation studies to validate the effectiveness of each component. Our contributions are:
\begin{itemize}
    \item We introduce LAW, a closed-loop VLM--WAM framework that unifies language reasoning, action generation, and world modeling through semantic subtask progress. The LAW agent coordinates planning, asynchronous completion assessment, and value-guided action fusion for long-horizon execution.
    \item We introduce MoTE, dual-stream hierarchical memory, and subtask-local value learning to support progress-aware control: MoTE strengthens joint video--action modeling through shared and modality-specific experts; dual-stream hierarchical memory preserves task-relevant history and subtask-boundary observations; and subtask-local value learning retains positive supervision from completed stages of failed trajectories.
    \item We demonstrate LAW's effectiveness on RoboTwin, RoboDojo, and long-horizon real-world manipulation through extensive experiments and ablation studies. Integrating OpenWAM-$\alpha$ achieves 95.56\% average success on RoboTwin, establishing a new open-source state of the art and demonstrating LAW's extensibility to stronger pretrained WAMs.
\end{itemize}

\section{Related Work}

\subsection{Vision-Language-Action Models for Robot Control}

Pretrained vision-language models (VLMs) provide semantic representations that vision-language-action (VLA) policies connect to executable control~\cite{driess2023palme,zitkovich2023rt2,li2024roboflamingo,kim2024openvla,black2024pizero}, supported by heterogeneous robot datasets and generalist pretraining~\cite{openx2023rtx,octo2024}. RT-2 and OpenVLA represent actions as tokens, whereas $\pi_0$ couples a VLM with a flow-matching expert for continuous action generation~\cite{zitkovich2023rt2,kim2024openvla,black2024pizero}. Beyond action parameterization, CoT-VLA predicts intermediate visual subgoals~\cite{zhao2025cotvla}, while VEGA aligns visual-encoder outputs with geometry-aware features before vision-language fusion~\cite{wang2026vega}. These approaches strengthen semantic transfer and spatial grounding, motivating complementary mechanisms for coordinating instructions, execution progress, and action selection over extended tasks.

Language-guided systems connect reasoning to execution through affordance-based skill selection, environmental feedback, action hierarchies, and spatial value maps~\cite{ahn2022saycan,huang2022innermonologue,belkhale2024rth,huang2023voxposer}. SayCan grounds language plans in learned affordances, whereas VoxPoser translates language-specified objectives and constraints into spatial maps for trajectory optimization~\cite{ahn2022saycan,huang2023voxposer}. In navigation, NavCoT structures decisions through future imagination and observation matching~\cite{lin2025navcot}, while Memoir retrieves environmental and behavioral experience using imagined states~\cite{xu2026memoir}. Although navigation differs from continuous manipulation, these studies illustrate how intermediate reasoning and selective memory access support sequential decisions. For long-horizon robot policies, the corresponding challenge is to retain task-relevant history and execution feedback when semantic stages extend beyond individual action chunks~\cite{shi2026memoryvla,torne2026mem}.

\subsection{World Action Models and Video-Based Policy}

World models support control through action-conditioned visual prediction or latent dynamics~\cite{ebert2018visual,hafner2020dream,zhou2024dinowm,assran2025vjepa2,bar2025nwm}, while video generators also enable synthetic robot training trajectories~\cite{jang2025dreamgen}. Visual Foresight optimizes actions through predicted observations, whereas Dreamer learns behavior through latent imagination, providing two complementary uses of predictive dynamics~\cite{ebert2018visual,hafner2020dream}. Recent world-action models (WAMs) connect future visual modeling with executable action generation through joint diffusion or future-latent alignment~\cite{zhu2025uwm,zheng2025flare,liang2025videopolicy,liao2026genie,zhang2026lingbotva,yuan2026fastwam,zhang2026lingbotva2}. Their designs differ in how predictive information is used during deployment: LingBot-VA combines causal video--action modeling with observation feedback and asynchronous execution~\cite{zhang2026lingbotva}, whereas Fast-WAM retains video co-training but omits test-time future prediction~\cite{yuan2026fastwam}. WAM4D further introduces geometric supervision through spatial register tokens and modality-specific causal attention~\cite{li2026wam4d}. These developments distinguish predictive supervision benefits from the cost of generating future observations online.

Reliable world-model-based control also requires faithful action consequences and informative evaluation signals. WorldEcho diagnoses failures to follow off-expert actions, and WorldSync improves action-conditioned generation through broader coverage and explicit action-effect alignment~\cite{chen2026worldsync}. Cosmos Policy jointly generates actions, future states, and values for planning~\cite{kim2026cosmospolicy}; MV-WAM integrates progress-value estimation with value-guided rollback~\cite{chen2026mvwam}; and SafeDojo optimizes VLA policies using imagined trajectories with separate task-reward and safety-cost estimates~\cite{tang2026safedojo}. LAW builds on these ingredients by coordinating semantic planning, asynchronous state assessment, and value-guided WAM candidate selection, supported by hierarchical visual memory. Its modality-specific and shared experts strengthen video--action interaction, while separate hierarchical memory streams preserve temporal context for language-level reasoning and continuous robot action generation. Unlike the trajectory-level returns used by MV-WAM~\cite{chen2026mvwam}, LAW terminates value targets at semantic subtask boundaries, preserving supervision from completed prefixes of unsuccessful episodes. This links local action evaluation to progress-aware action selection and planning grounded in semantic subtasks.

\section{Method}

\subsection{LAW Framework Overview}

\begin{figure*}[!tb]
    \centering
    \includegraphics[width=\textwidth]{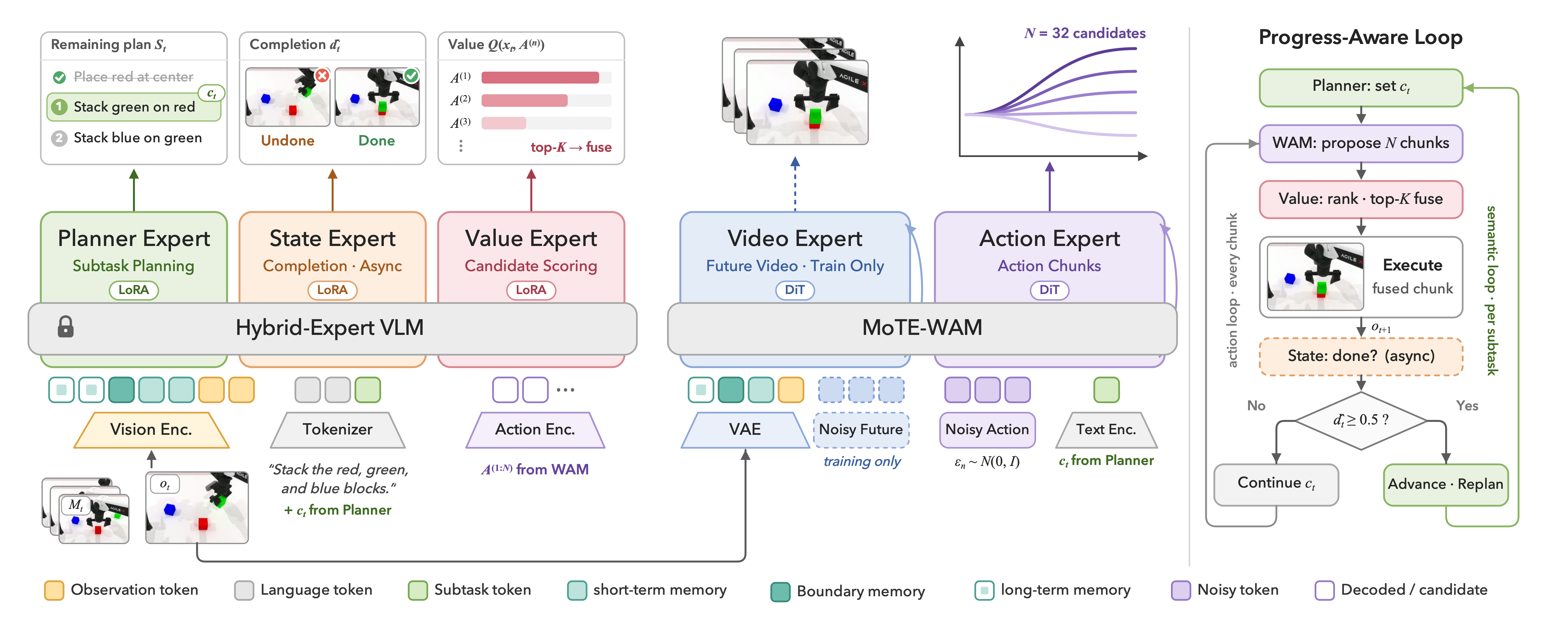}
    \caption{LAW architecture and progress-aware control loop. Planner, state, and value experts share a frozen VLM backbone. MoTE-WAM jointly learns video and action prediction, with future video prediction used only during training. At inference, the value expert ranks and fuses candidate action chunks, while asynchronous completion assessment determines whether to continue the current subtask or advance and replan.}
    \label{fig:law-framework}
\end{figure*}

Figure~\ref{fig:law-framework} provides an overview of the LAW agent's architecture and loop. The LAW agent factorizes long-horizon control into a progress-aware semantic control loop and a short-horizon action loop. At control step $t$, the remaining plan uses local indices, with its first element $c_t$ defining the current subtask, while two memory streams, $\mathcal M_t^{\mathrm V}$ and $\mathcal M_t^{\mathrm W}$, provide task-relevant history to the VLM and WAM, respectively. The current subtask connects the two loops: it specifies the objective for action generation and value estimation, while execution feedback determines whether that objective should be continued or advanced.

We write $o_t=(I_t,p_t)$ for the available multiview RGB images and proprioceptive state. The WAM uses RGB, EEF state, and visual memory; the VLM uses RGB and visual memory, with joint-state values additionally supplied to the state expert. Each module selects its required components of $o_t$, and visual-memory encoders operate on $I_t$. Input and action conventions are detailed in Appendix~\ref{app:observation_action}.

Given a global instruction $g$, current observation $o_t$, and VLM memory $\mathcal M_t^{\mathrm V}$, the planner predicts the remaining semantic subtask sequence
\begin{equation}
    \mathcal{S}_t
    = \mathrm{VLM}_{\mathrm{plan}}(g,o_t,\mathcal{M}_t^{\mathrm V})
    = (s_{t,1},s_{t,2},\ldots,s_{t,L_t}),
    \label{eq:law_subtasks}
\end{equation}
where $L_t$ is the number of remaining subtasks and $c_t=s_{t,1}$. Each newly predicted plan is indexed locally from one. An independent, monotonically increasing subtask-instance identifier $u_t$ associates asynchronous predictions with the current execution instance. The planner is invoked at episode initialization and at semantic decision events rather than every control step. Conditioned on $c_t$, the current observation, and WAM memory, the WAM samples $N$ candidate action chunks:
\begin{equation}
    \big\{A_t^{(n)}\big\}_{n=1}^{N}
    = \mathrm{WAM}_{\mathrm{act}}(o_t,c_t,\mathcal{M}_t^{\mathrm W};\epsilon_n).
    \label{eq:law_action}
\end{equation}
Each candidate $A_t^{(n)}=\{a_{t:t+H-1}^{(n)}\}$ contains an action sequence of horizon $H=32$, and $\epsilon_n$ is an independent noise initialization that induces proposal diversity. Following Fast-WAM~\cite{yuan2026fastwam}, the video branch provides a joint video--action training signal but is not sampled to produce explicit future images at inference. With VLM context
\begin{equation}
    x_t=(g,o_t,c_t,\mathcal M_t^{\mathrm V}),
    \label{eq:law_context}
\end{equation}
the value expert scores candidate action chunks and fuses the top-$K$ proposals for execution:
\begin{equation}
    q_t^{(n)}=Q_\theta(x_t,A_t^{(n)}),
    \qquad
    \widehat A_t
    =\operatorname{Fuse}_{K,T}
    \!\left(\{A_t^{(n)},q_t^{(n)}\}_{n=1}^{N}\right),
    \label{eq:law_select}
\end{equation}
Here, $Q_\theta$ denotes the subtask-local value function implemented by the value expert, which estimates the expected return of a candidate under the current subtask. We use $K=8$, and $T$ is the fusion temperature. The robot executes a prefix of $\widehat A_t$ before the WAM generates new proposals from the next observation. Section~\ref{sec:value_learning} specifies the subtask-local target and the fusion operator.

In parallel with action execution, the state expert estimates whether the current subtask has completed:
\begin{equation}
    \hat d_t
    =\mathrm{VLM}_{\mathrm{state}}
    (g,o_t,c_t,\mathcal M_t^{\mathrm V}).
    \label{eq:law_state}
\end{equation}
The index $t$ in the completion prediction refers to the acquisition time of its input observation, not the inference completion time. Each prediction carries this timestamp and the corresponding subtask identifier. The completion prediction is binarized at $0.5$. In the current implementation, subtask completion determines the semantic control event
\begin{equation}
e_t=
\begin{cases}
    \text{\textsc{Advance}}, & \hat d_t\geq 0.5,\\
    \text{\textsc{Continue}}, & \text{otherwise}.
\end{cases}
\label{eq:law_event}
\end{equation}
A \textsc{Continue} event preserves $c_t$, $u_t$, and the current plan. An \textsc{Advance} event stores the observation at the control-loop transition as a subtask-boundary state and terminates the episode if the final subtask has completed. Otherwise, the planner predicts the remaining subtasks again using the updated observation and memory, and the LAW agent switches to the next predicted subtask. The new remaining sequence is indexed locally from one, its first element becomes $c_{t+1}$, and the execution-instance identifier advances to $u_{t+1}=u_t+1$. Completion of the sole remaining subtask terminates the episode. Thus, semantic subtask completion changes the conditioning objective, memory state, and planning schedule of subsequent action generation. This feedback implements the temporal interface introduced above. The following subsections detail the VLM experts, MoTE, memory, and subtask-local value learning.

Algorithm~\ref{alg:law-inference} summarizes the complete inference loop. Newly acquired observations trigger asynchronous completion assessment using the latest observation, without queuing historical frames. The control loop reads available results without blocking action execution. A result is valid only if its subtask-instance identifier matches $u_t$ and its observation timestamp is newer than that of the last consumed result; results superseded by newer available predictions are discarded. If no valid result is available, execution continues under the current subtask and plan. Upon accepting completion, the agent stores $o_{t+1}$ at the control-loop transition as the boundary observation. This observation marks the actual subtask switch rather than the earlier observation on which completion was assessed.

\begin{algorithm}[t]
    \caption{Progress-aware inference of the LAW agent}
    \label{alg:law-inference}
    \begin{algorithmic}[1]
        \Require Instruction $g$, initial observation $o_0$, candidate count $N$, top-$K$ size $K$, temperature $T$
        \State Initialize $t\gets0$, $\mathcal M_0^{\mathrm V}\gets\emptyset$, and $\mathcal M_0^{\mathrm W}\gets\emptyset$
        \State $\mathcal S_0\gets\mathrm{VLM}_{\mathrm{plan}}(g,o_0,\mathcal M_0^{\mathrm V})$ , $c_0\gets s_{0,1}$, and $u_0\gets0$
        \While{the task is not terminated}
            \State $\{A_t^{(n)}\}_{n=1}^{N}\gets\mathrm{WAM}_{\mathrm{act}}(o_t,c_t,\mathcal M_t^{\mathrm W};\epsilon_n)$
            \State $x_t\gets(g,o_t,c_t,\mathcal M_t^{\mathrm V})$
            \For{$n=1,\ldots,N$}
                \State $q_t^{(n)}\gets Q_\theta(x_t,A_t^{(n)})$
            \EndFor
            \State $\widehat A_t\gets\operatorname{Fuse}_{K,T}(\{A_t^{(n)},q_t^{(n)}\}_{n=1}^{N})$
            \State Execute a prefix of $\widehat A_t$; assess each new observation asynchronously using its timestamp and $u_t$
            \State Receive $o_{t+1}$ and update $\mathcal M_{t+1}^{\mathrm V}$ and $\mathcal M_{t+1}^{\mathrm W}$
            \State $c_{t+1}\gets c_t$, $u_{t+1}\gets u_t$, and $\mathcal S_{t+1}\gets\mathcal S_t$
            \State Non-blockingly read the newest unconsumed result $(\hat d,\tau,u)$ for $u_t$; discard superseded results
            \If{a result newer than the last consumed timestamp with $u=u_t$ is available and $\hat d\geq0.5$}
                \State Store $o_{t+1}$ at the control-loop transition as a subtask-boundary observation
                \If{$|\mathcal S_t|=1$}
                    \State \textbf{terminate successfully}
                \Else
                    \State $u_{t+1}\gets u_t+1$
                    \State $\mathcal S_{t+1}\gets\mathrm{VLM}_{\mathrm{plan}}(g,o_{t+1},\mathcal M_{t+1}^{\mathrm V})$
                    \State Set $c_{t+1}$ to the first subtask of $\mathcal S_{t+1}$
                \EndIf
            \Else
                \State $c_{t+1}\gets c_t$ and $u_{t+1}\gets u_t$
            \EndIf
            \State Mark any valid result read in this iteration as consumed
            \State $t\gets t+1$
        \EndWhile
    \end{algorithmic}
\end{algorithm}

\begin{figure*}[!t]
    \centering
    \includegraphics[width=0.99\textwidth]{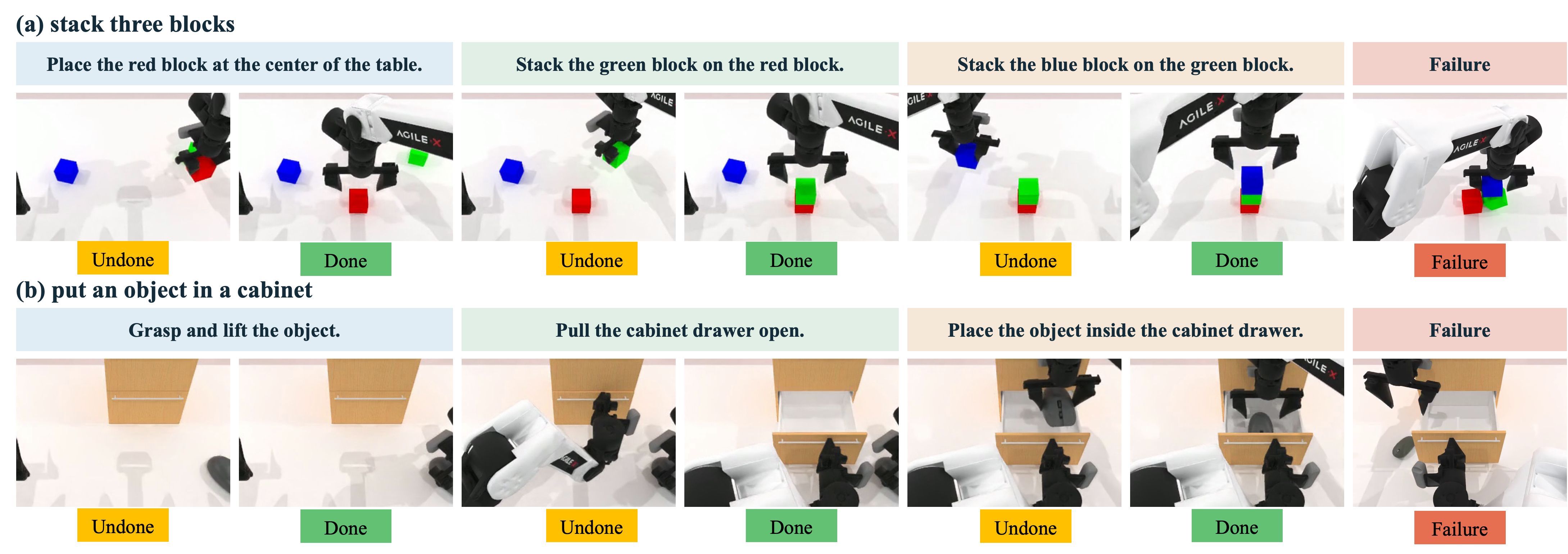}
    \caption{Visualization of semantic subtask annotations, temporal boundaries, and failure cases in RoboTwin.}
    \label{fig:subtask}
\end{figure*}

\subsection{Hybrid-Expert VLM for Progress-Aware Control}

\noindent\textbf{Hybrid-Expert VLM}
To support planning, execution monitoring, and action evaluation within a single VLM, we introduce a hybrid-expert architecture with task-specific LoRA adapters~\cite{hu2021lora}. The planner, state, and value experts share a frozen VLM backbone, allowing them to reuse common visual-language representations while avoiding the cost of maintaining a separate full model for each control function. Each expert activates its own adapter and task-specific head, enabling specialization under distinct supervision objectives while isolating parameter updates and reducing interference across functions. Expert adapters are selected by control function rather than token-level routing.

The planner is trained with semantic subtask supervision. We construct ordered subtask annotations and temporal boundaries from the demonstrations, as illustrated in Fig.~\ref{fig:subtask}. Additional task decompositions and segmented examples are provided in Appendix~\ref{app:subtask_decomposition}. Given the global instruction, visual observation, and execution memory, the planner is trained to generate the remaining subtask sequence autoregressively:
\begin{equation}
    \mathcal{L}_{\mathrm{plan}}
    =-\sum_{i=1}^{|\mathcal{S}|}
    \log p_\theta\!\left(s_i\mid g,o_t,\mathcal{M}_t^{\mathrm V},s_{<i}\right).
    \label{eq:plan_loss}
\end{equation}

The state expert is trained with subtask completion supervision and an auxiliary failure-prediction objective to improve completion assessment. Completion labels are derived from motion-based progress within annotated subtask intervals; the label construction is detailed in Appendix~\ref{app:progress_labels}. Failure trajectories are additionally reviewed to identify the first visually unambiguous manipulation error, together with its semantic stage and visual evidence. During training, dedicated completion and failure query tokens provide complementary supervision: the auxiliary failure head encourages the shared representation to distinguish unsuccessful execution from completed subtasks. The training objective is
\begin{equation}
    \mathcal{L}_{\mathrm{state}}
    =\lambda_d\,\mathrm{BCE}(\hat d_t,d_t)
    +\lambda_f\,\mathrm{BCE}(\hat f_t,f_t).
    \label{eq:state_loss}
\end{equation}
where $d_t$ and $f_t$ are binary completion and failure labels, and $\hat d_t$ and $\hat f_t$ are the corresponding training predictions. The auxiliary failure loss improves the features used by the completion head. For value prediction, candidate action chunks $A_t$ are encoded as tokens and processed jointly with the context $x_t$ by the shared VLM. A learned query token aggregates these features for a 101-bin categorical value head, yielding
\begin{equation}
    Q_\theta(x_t,A_t)=\sum_{b=1}^{101}z_b\,p_\theta(b\mid x_t,A_t),
    \label{eq:law_value_prediction}
\end{equation}
where $p_\theta$ is the softmax output and $z_b$ are fixed bin centers. The value expert uses a LoRA adapter~\cite{hu2021lora} initialized from the state expert; only this adapter, the action encoder, query embedding, and value head are trained, while the shared backbone, existing experts, and WAM remain frozen.

\begin{figure*}[!tb]
    \centering
    \includegraphics[width=0.99\textwidth]{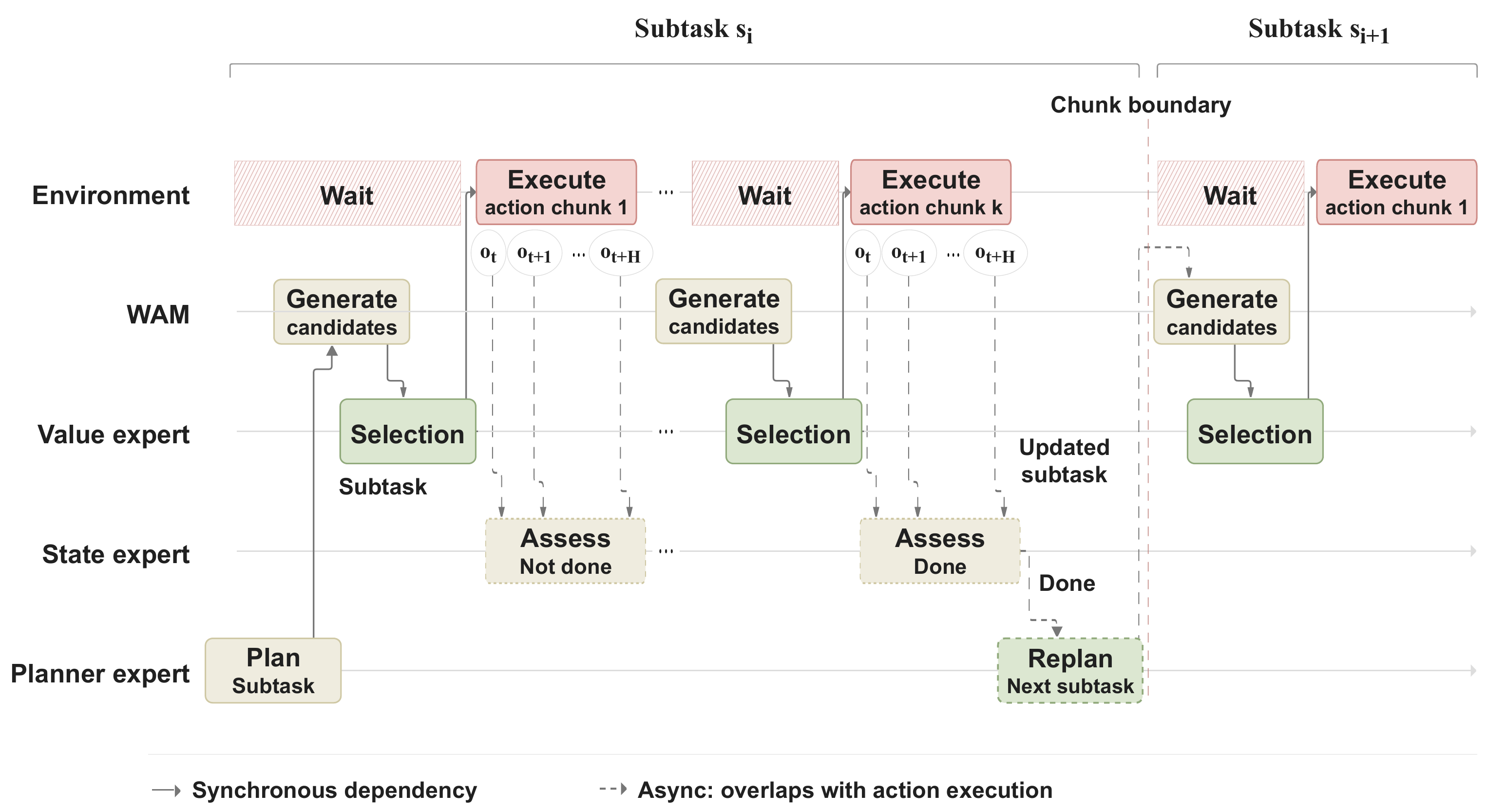}
    \caption{Mixed-frequency coordination for progress-aware control in LAW. The planner operates at semantic subtask boundaries, while WAM generation and value-based selection synchronously determine each action chunk. The state expert asynchronously estimates subtask completion during action execution and triggers the next planning decision when the current subtask is completed.}
    \label{fig:mixed-frequency-coordination}
\end{figure*}

\noindent\textbf{Mixed-Frequency Coordination}
The three experts and the WAM operate at different frequencies, as illustrated in Fig.~\ref{fig:mixed-frequency-coordination}. The planner is invoked at episode initialization and near subtask boundaries to produce or update the remaining plan. After the WAM generates candidate action chunks, the value expert synchronously scores and fuses them into the action chunk executed by the robot. In contrast, the state expert runs asynchronously during action execution. Each query-based state-expert forward pass takes approximately $60$ ms and runs concurrently with action execution, enabling non-blocking completion assessment on the latest observations. The resulting pipelined parallelism between action execution and state assessment improves inference throughput by $15\%$. A valid prediction with completion probability at least $0.5$ triggers a subtask switch at a control-loop boundary; validity is determined by the timestamp and subtask identifier as described in Algorithm~\ref{alg:law-inference}.

\subsection{Joint Video--Action Modeling with MoTE}

The WAM must jointly model the evolution of visual observations and the robot actions that cause these changes. In the original Mixture-of-Transformers (MoT)~\cite{liang2025mot}, video and action branches exchange information through mixed attention: their queries, keys, and values are concatenated for joint attention, after which the outputs are returned to the two modality-specific branches~\cite{zhang2026lingbotva,yuan2026fastwam}. Mixed attention exchanges information, but modality-specific transformations limit shared video--action modeling.

\begin{figure*}[!tb]
    \centering
    \includegraphics[width=0.99\textwidth]{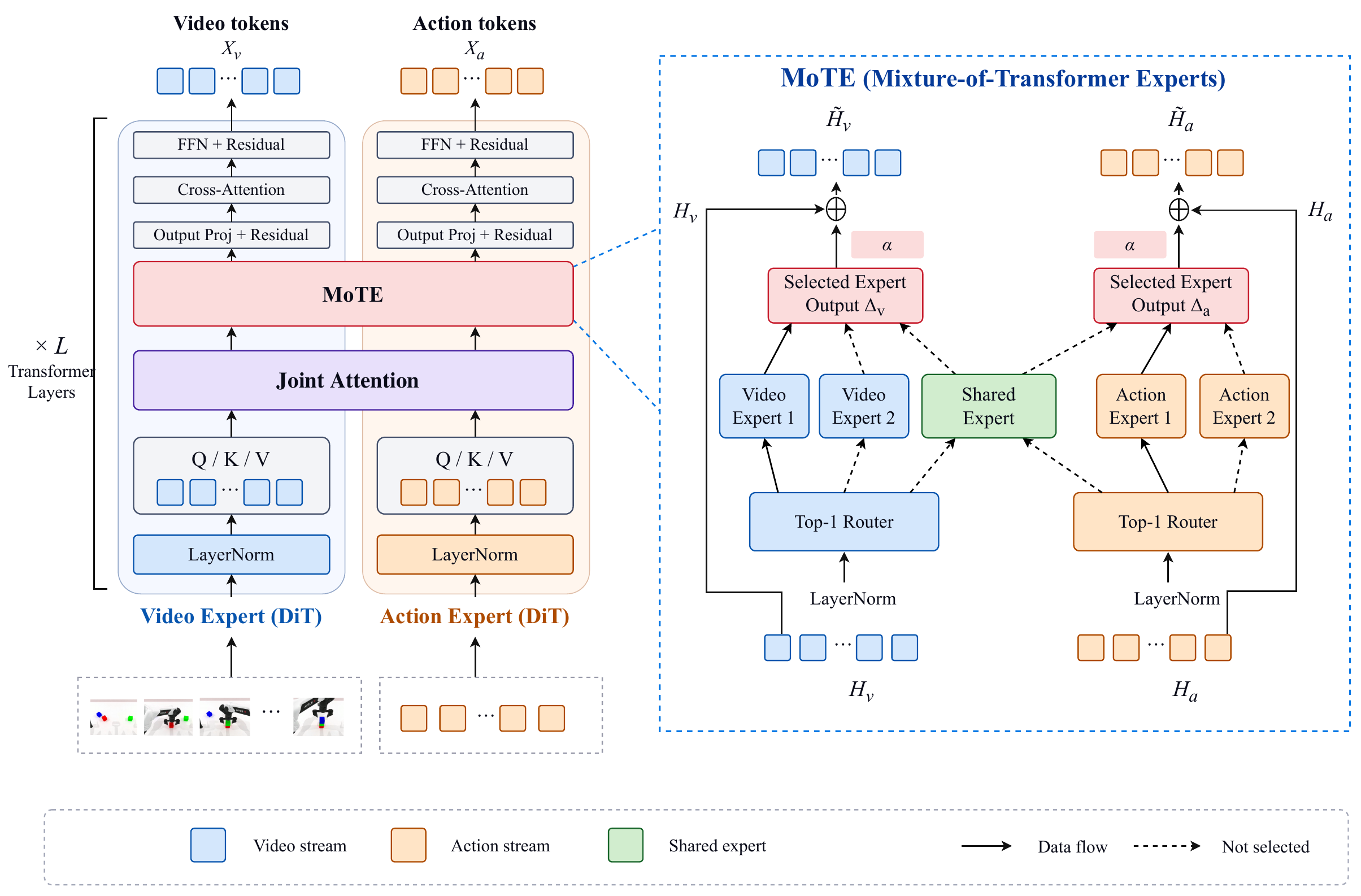}
    \caption{MoTE architecture for joint video--action modeling. The video and action DiT branches exchange information through MoT mixed self-attention. After mixed attention, modality-specific top-1 routers select private experts or a shared expert available to both streams. The routed expert output is scaled by $\alpha$ and added residually before the remaining Transformer transformations.}
    \label{fig:mote-architecture}
\end{figure*}

We introduce Mixture-of-Transformer Experts (MoTE) to strengthen this interaction, as illustrated in Fig.~\ref{fig:mote-architecture}. MoTE applies a residual mixture-of-experts layer to the mixed-attention representation. The router produces sparse expert weights for the current modality:
\begin{equation}
    \mathbf{w}
    =\operatorname{TopKSoftmax}\!\left(W_r\operatorname{LN}(\mathbf{H})\right),
    \label{eq:mote_router}
\end{equation}
where $\mathbf{H}$ is the mixed-attention representation and $W_r$ is a modality-specific router. The routed residual update is
\begin{equation}
    \widetilde{\mathbf{H}}
    =\mathbf{H}
    +\alpha\sum_{e\in\mathcal{E}}w_e E_e\!\left(\operatorname{LN}(\mathbf{H})\right),
    \label{eq:mote_update}
\end{equation}
where $\mathcal{E}$ denotes the selected experts, $w_e$ is the routing weight, and $E_e$ is an expert network. The modality-specific experts preserve the distinct statistics of video and action tokens, while the shared experts capture their common temporal and physical dynamics. Because the MoTE layer is applied immediately after mixed attention, each token is routed to either a modality-private expert or the shared expert before entering the next Transformer block. The shared expert is available to both modalities; top-1 routing does not apply private and shared experts simultaneously to the same token.

In the default configuration, each modality has two private experts and both modalities share one additional expert. The router uses top-1 selection and the residual scale is set to $0.1$. The training objective combines video prediction, action prediction, and router regularization:
\begin{equation}
    \mathcal{L}_{\mathrm{WAM}}
    =\lambda_v\mathcal{L}_{\mathrm{video}}
    +\lambda_a\mathcal{L}_{\mathrm{action}}
    +\lambda_{\mathrm{bal}}\mathcal{L}_{\mathrm{balance}}
    +\lambda_z\mathcal{L}_{z}.
    \label{eq:mote_loss}
\end{equation}
Here, $\mathcal{L}_{\mathrm{balance}}$ encourages balanced expert usage and $\mathcal{L}_{z}$ regularizes the router logits. We set $\lambda_v=\lambda_a=1$ and $\lambda_{\mathrm{bal}}=\lambda_z=10^{-4}$ in the default WAM configuration. MoTE supplements mixed attention with modality-specific and shared nonlinear transformations to jointly model scene evolution and actions.

\subsection{Dual-Stream Hierarchical Memory}
\label{sec:hierarchical_memory}

Long-horizon manipulation requires historical context to assess task progress and maintain consistent action generation~\cite{shi2026memoryvla,torne2026mem}. Retaining every past observation is computationally expensive, while compressing the entire history into a single representation can discard recent motion details and critical subtask transitions. LAW maintains separate VLM and WAM memory streams over shared observations.

\begin{figure*}[!t]
    \centering
    \includegraphics[width=\textwidth]{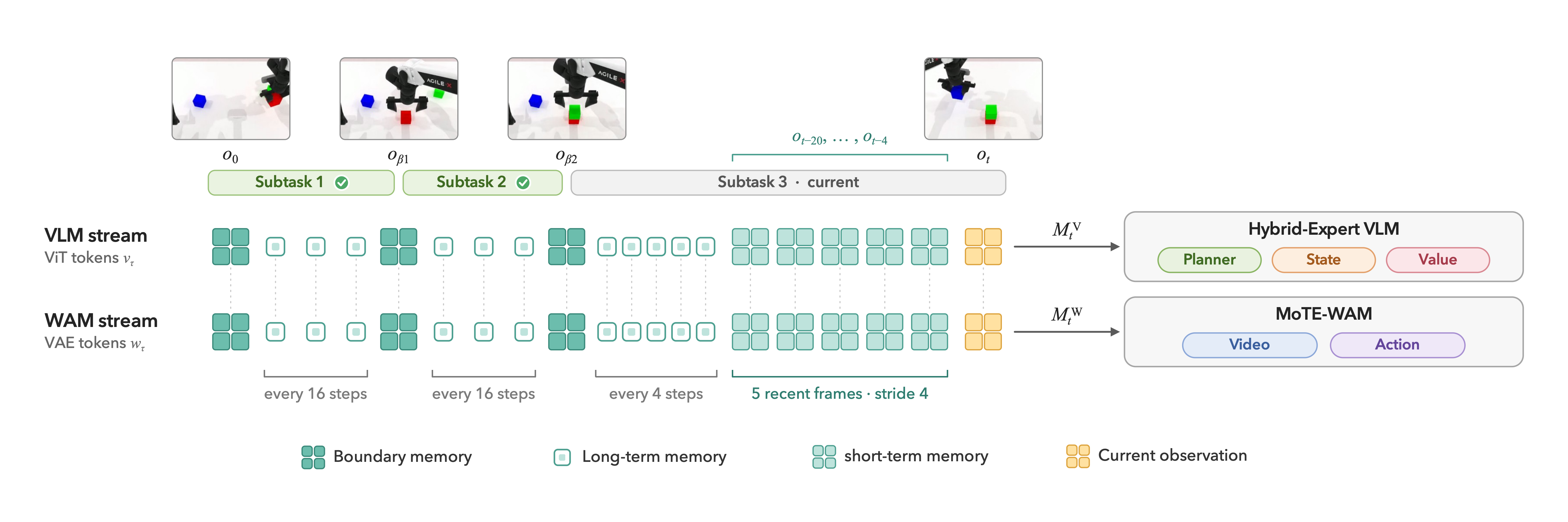}
    \caption{Dual-stream hierarchical visual memory. VLM and WAM streams encode the same observation history as ViT and VAE tokens, respectively. Short-term memory and boundary memory retain full-resolution visual tokens, while long-term memory stores spatially compressed tokens.}
    \label{fig:hierarchical-memory-attention}
\end{figure*}

At time $t$, we retain the previous five observations at a stride of four, giving the recent window $\{o_{t-20},\allowbreak o_{t-16},\allowbreak o_{t-12},\allowbreak o_{t-8},\allowbreak o_{t-4}\}$; unavailable elements are omitted near the beginning of an episode, and the current observation $o_t$ is provided separately. We also retain the initial observation and observations recorded at control-loop subtask transitions when they fall outside this recent window. Earlier frames are sampled every four steps within the current subtask and every 16 steps within completed subtasks. All selected timestamps share the phase $t\bmod4$, and a subtask-boundary frame is stored only once. Let $\mathcal C_t=(c_1,\ldots,c_m)$ denote the resulting long-term timestamps and $\mathcal B_t=(\beta_1,\ldots,\beta_p)$ the retained boundary timestamps, both in temporal order. Figure~\ref{fig:hierarchical-memory-attention} illustrates how the VLM and WAM streams organize the same short-term, subtask-boundary, and long-term observation history in their respective token spaces.

The two streams encode these observations in their native token spaces. Let $\mathbf v_\tau=V(o_\tau)$ be the VLM visual tokens and $\mathbf w_\tau=W(o_\tau)$ the WAM VAE tokens. Both the VLM and WAM compress long-term memory using spatial average pooling, yielding one summary token per older observation, denoted by $\bar{\mathbf v}_\tau$ and $\bar{\mathbf w}_\tau$, respectively. Boundary and recent observations retain their full spatial token sequences.

Let $\mathbf x_\tau^X$ and $\bar{\mathbf x}_\tau^X$ denote the full and compressed tokens of stream $X$, respectively, where $(\mathbf x^{\mathrm V},\bar{\mathbf x}^{\mathrm V})=(\mathbf v,\bar{\mathbf v})$ and $(\mathbf x^{\mathrm W},\bar{\mathbf x}^{\mathrm W})=(\mathbf w,\bar{\mathbf w})$. Using the same timestamps, both memory streams can then be written as
\begin{equation}
\begin{aligned}
    \mathcal M_t^X
    =\Big[&\{\bar{\mathbf x}_{c_i}^{X}\}_{i=1}^{m};\{\mathbf x_{\beta_i}^{X}\}_{i=1}^{p};\{\mathbf x_{t-4j}^{X}\}_{j=1}^{5}\Big],
            X\in\{\mathrm V,\mathrm W\}.
\end{aligned}
\label{eq:dual_stream_memory}
\end{equation}
The three groups correspond to compressed long-term history, full-resolution subtask boundaries, and full-resolution recent observations. Thus, $\mathcal M_t^{\mathrm V}$ supports planning, state estimation, and candidate evaluation, while $\mathcal M_t^{\mathrm W}$ conditions video prediction and action generation.

\subsection{Subtask-Local Value Learning}
\label{sec:value_learning}


\begin{figure}[!tb]
    \centering
    \includegraphics[width=\columnwidth]{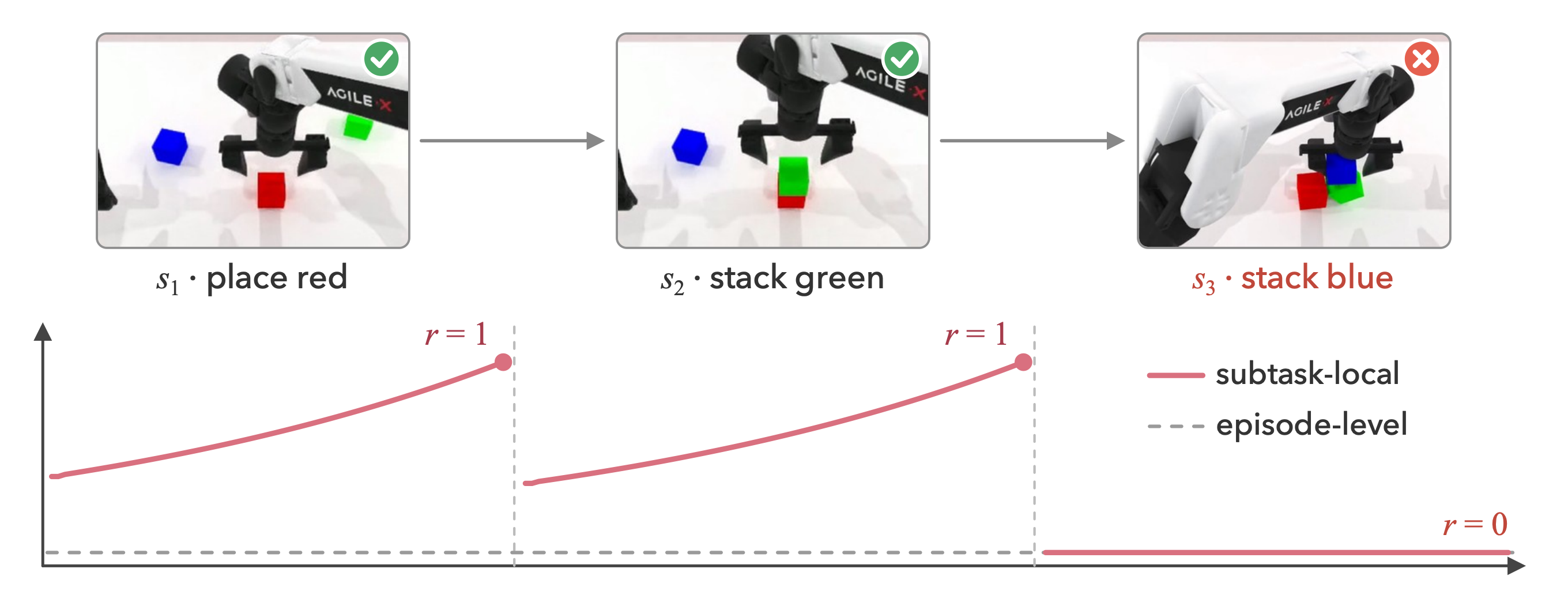}
    \caption{Subtask-local value supervision retains positive targets for completed subtasks even when a later subtask fails, whereas episode-level supervision assigns zero return to the failed episode.}
    \label{fig:subtask-local-value}
\end{figure}

\noindent\textbf{Subtask-local value supervision.}
Episode-level outcomes discard useful supervision when a trajectory completes several subtasks before failing. Figure~\ref{fig:subtask-local-value} illustrates how subtask-local supervision preserves credit for completed stages of a failed episode. We instead assign reward $1$ to the terminal action of each successfully completed subtask and $0$ elsewhere. Completed prefixes therefore retain positive supervision, while the final failed subtask receives no positive reward and unobserved subtasks are excluded. For a chunk with $h_t$ valid actions, the TD target is
\begin{equation}
    y_t=\sum_{i=0}^{h_t-1}\gamma^i r_{t+i}
    +b_t\gamma^{h_t}Q_{\bar\theta}(x_{t+h_t},A^{\mathrm{data}}_{t+h_t}),
    \label{eq:value_td}
\end{equation}
where $A^{\mathrm{data}}_{t+h_t}$ is the next recorded chunk and $\bar\theta$ denotes exponential-moving-average target parameters. Each chunk is truncated at the first subtask boundary, and $h_t$ counts only its valid actions; actions beyond that boundary are masked out of the target. We set $b_t=1$ only when the next recorded chunk exists within the same subtask and no terminal action is included; otherwise, $b_t=0$. Thus, terminal chunks and chunks without a valid within-subtask successor do not bootstrap.

Building on categorical value learning~\cite{farebrother2024stop}, we clip $y_t$ to the value support, obtaining $\widetilde y_t$, and construct soft labels using a normalized Gaussian kernel evaluated at the bin centers. The value expert is trained with categorical cross-entropy:
\begin{equation}
    \begin{aligned}
        \rho_{t,b}
        &\propto\exp\!\left[-\frac{(z_b-\widetilde y_t)^2}{2\sigma^2}\right],\\
        \mathcal L_{\mathrm{value}}
        &=-\mathbb E_{\mathcal D}\!\left[
        \sum_{b=1}^{101}\rho_{t,b}\log p_\theta(b\mid x_t,A_t)\right],
    \end{aligned}
    \label{eq:value_loss}
\end{equation}
where $\rho_t$ is normalized over bins and detached.

\noindent\textbf{Value-guided action selection.}
At execution time, the WAM generates candidate chunks $\{A_t^{(n)}\}_{n=1}^{N}$. The value expert retains the top-$K$ index set $\mathcal E_t$ and assigns temperature-controlled weights
\begin{equation}
    w_n=
    \frac{\exp\!\left(Q_\theta(x_t,A_t^{(n)})/T\right)}
    {\sum_{k\in\mathcal E_t}\exp\!\left(Q_\theta(x_t,A_t^{(k)})/T\right)},
    \qquad n\in\mathcal E_t,
    \label{eq:value_weights}
\end{equation}
where $T>0$ is the temperature. Each weight is shared across the candidate's time steps. Positions and gripper commands are averaged linearly. For each arm and time step, normalized candidate quaternions $q_n$ are combined using the weighted Markley mean~\cite{markley2007averaging}:
\begin{equation}
    M=\sum_{n\in\mathcal E_t}w_n q_nq_n^\top,
    \qquad
    \widehat q\in\underset{\|q\|_2=1}{\operatorname{arg\,max}}\;q^\top Mq.
    \label{eq:value_markley}
\end{equation}
The aggregate is a unit eigenvector of $M$ associated with its largest eigenvalue, invariant as a rotation to candidate sign flips $q_n\mapsto-q_n$. The robot executes the aggregated chunk prefix, then repeats with new observations.

\section{Experiments}

\subsection{Training Details}
\label{sec:training_details}

\noindent\textbf{Training datasets.}
We use task-specific training data from RoboTwin~\cite{chen2025robotwin}, RoboDojo~\cite{chen2026robodojo}, and real-world manipulation. RoboTwin covers 50 bimanual manipulation tasks; its public Aloha-AgileX release provides 50 clean and 500 domain-randomized demonstrations per task. The official RoboDojo training release contains 3,500 trajectories across 35 task directories, with 100 trajectories per directory collected through automated synthesis or VR-based teleoperation. We augment the simulation trajectories with semantic subtask annotations and temporal boundaries for planning and completion assessment.

For real-world training, we collect 300 teleoperated trajectories per task on the AgileX Cobot Magic platform, totaling 900 trajectories, with one scene camera and two wrist cameras. Puzzle comprises three subtasks, each inserting one colored piece into its matching recess. Domino comprises five subtasks: standing four dominoes individually and then toppling the row. Organize comprises nine subtasks: opening the corresponding drawer, placing a colored block inside, and closing the drawer, repeated for three blocks. Each trajectory includes multiview observations, robot states, actions, language instructions, and semantic subtask boundaries. Successful and failed trajectories provide complementary supervision for value learning. Additional annotation and collection details are given in Appendices~\ref{app:subtask_decomposition} and~\ref{app:real_world_data}.

\noindent\textbf{Overall training pipeline.}
LAW is trained in two stages. In the first stage, we train the planner and state experts sequentially on semantic subtask annotations, while training the WAM independently on paired video and robot-action sequences. In the second stage, we freeze these components and train the value expert on successful demonstrations and annotated failure trajectories using the subtask-local objective in Section~\ref{sec:value_learning}. All parameters are frozen for evaluation. This procedure prevents value learning from destabilizing the generator, planner, or state expert.

\begin{table}[!tb]
    \centering
    \caption{Training hyperparameters of LAW.}
    \label{tab:training_hyperparameters}
    
    \begin{adjustbox}{max width=\columnwidth,max totalheight=0.88\textheight}
    \lawtablefont
\begin{tabular}{@{}lccccc@{}}
        \toprule
        \textbf{Stage} & \textbf{Steps} & \textbf{LR} & \textbf{Batch} & \textbf{Acc.} & \textbf{GPUs} \\
        \midrule
        Planner expert & 3,000 & $1\!\times\!10^{-5}$ & 6 & 1 & 16 \\
        State expert & 50,000 & $1\!\times\!10^{-5}$ & 6 & 1 & 16 \\
        WAM & 30,000 & $1\!\times\!10^{-4}$ & 6 & 2 & 32 \\
        Value expert & 50,000 & $1\!\times\!10^{-5}$ & 8 & 1 & 16 \\
        \bottomrule
    \end{tabular}
    \end{adjustbox}
    
\end{table}

\noindent\textbf{Hybrid-Expert VLM.}
We use Qwen3-VL-2B-Instruct~\cite{bai2025qwen3} as the shared VLM backbone and keep its language and vision parameters frozen. The planner and state adapters use rank $r=64$, while the value adapter uses rank $r=128$, adding approximately 100M parameters in total. The planner and state adapters use LoRA scale 128 and dropout 0.05. After training the planner, we freeze its adapter and train the state adapter together with its completion and auxiliary failure heads.

The value adapter is initialized from the state adapter, while the action encoder, value query, and categorical head are newly initialized. Each input contains a 32-step, 16-dimensional end-effector action chunk sampled at $0.06$ seconds per step. The value expert is trained exclusively on recorded action chunks from successful demonstrations and constructed failure trajectories, without using WAM-sampled candidates. Failure trajectories comprise 25\% of the training samples. Value training uses AdamW with zero weight decay, 1,000 warm-up steps followed by linear decay, and gradient clipping at 1.0. The categorical critic uses 101 bins over $[-0.01,1.01]$, discount factor $\gamma=0.99$, HL-Gauss width $\sigma=0.0075$, and an exponential-moving-average target with update rate $\tau=0.005$.

\noindent\textbf{Failure data construction.}
We construct failure trajectories in simulation by replaying successful demonstrations and injecting task-specific perturbations at selected manipulation stages. Each task is assigned one to three applicable failure modes, including missed grasps, premature releases, incorrect placements, missed contacts, and incomplete or misaligned task-specific operations. For each task--failure-type pair, we target 35 clean trajectories and 100 domain-randomized trajectories. We then filter out semantically invalid, visually ambiguous, and low-quality samples.

\noindent\textbf{World-action model.}
The WAM is initialized from Wan2.2-TI2V-5B~\cite{wan2025wan22ti2v5b}, with the action branch initialized from the pretrained ActionDiT weights~\cite{yuan2026fastwam}. We optimize the video and action DiTs, including the MoTE layers, while keeping the remaining pretrained components frozen. Training uses three camera views at $384\times384$ resolution, an action horizon of 32, and subtask-conditioned text embeddings. We use AdamW with $\beta_1=0.9$, $\beta_2=0.95$, weight decay $10^{-2}$, and gradient clipping at 1.0. The learning rate is linearly warmed up for the first 5\% of training and then follows a cosine schedule.

\noindent\textbf{Testing.}
The WAM samples 32 candidate chunks using three denoising steps, without explicitly generating future images. The value expert retains the top eight candidates with temperature $T=1$, and the robot executes their fused prefix before resampling from updated observations.

Table~\ref{tab:training_hyperparameters} lists the optimization settings; ``Batch'' is the per-GPU batch size and ``Acc.'' denotes gradient accumulation.

\subsection{Main Results}
\label{sec:main-results}

We evaluate LAW on 50 RoboTwin tasks under clean and randomized settings, comparing it with representative mainstream robot policies. As shown in Table~\ref{tab:robotwin-main}, LAW achieves success rates of 94.22\% and 94.20\%, respectively, yielding an average of 94.21\% and outperforming the selected baselines in both settings. Without robot-data pretraining, LAW improves over Fast-WAM (91.83\%) by 2.38 percentage points and exceeds LingBot-VA 2.0 (93.6\%), the best pretrained baseline in this comparison.

\begin{table}[!tb]
    \centering
    \caption{Comparison with mainstream robot policies on RoboTwin.}
    \label{tab:robotwin-main}
    \renewcommand{\arraystretch}{1.25}
    \setlength{\tabcolsep}{4pt}
    \setlength{\arrayrulewidth}{0.55pt}
    \begin{adjustbox}{max width=\columnwidth,max totalheight=0.88\textheight}
    \lawtablefont
\begin{tabular}{@{}l|ccc@{}}
        \toprule
        \textbf{Method} & \textbf{Clean (\%)} & \textbf{Rand. (\%)} & \textbf{Avg. (\%)} \\
        \midrule
        X-VLA~\cite{zheng2025xvla}$^{*}$ & 72.9 & 72.8 & 72.9 \\
        $\pi_{0.5}$~\cite{black2025pi05}$^{*}$ & 82.7 & 76.8 & 79.8 \\
        Qwen-RobotManip~\cite{yuan2026qwenrobotmanip}$^{*}$ & 93.4 & 92.5 & 93.0 \\
        Motus~\cite{bi2026motus}$^{*}$ & 88.7 & 87.0 & 87.9 \\
        Fast-WAM~\cite{yuan2026fastwam} & 91.88 & 91.78 & 91.83 \\
        LingBot-VA~\cite{zhang2026lingbotva}$^{*}$ & 92.9 & 91.5 & 92.2 \\
        LingBot-VA 2.0~\cite{zhang2026lingbotva2}$^{*}$ & 93.8 & 93.4 & 93.6 \\
        \midrule
        \textbf{LAW + Fast-WAM} & \textbf{94.22} & \textbf{94.20} & \textbf{94.21} \\
        \bottomrule
    \end{tabular}
    \end{adjustbox}%

\par\vspace{2pt}
    {\footnotesize\raggedright $^{*}$ denotes robot-data pretraining.\par}
\end{table}

\begin{table*}[!tb]
    \centering
    \caption{Comparison across RoboDojo capability dimensions.Results are averaged over three seeds.  }
    \label{tab:robodojo-main}
    \renewcommand{\arraystretch}{1.20}
    \setlength{\tabcolsep}{3.2pt}
    \begin{adjustbox}{max width=\textwidth,max totalheight=0.88\textheight}
    \lawtablefont
\begin{tabular}{@{}lccccccc@{}}
        \toprule
        \multicolumn{2}{c}{} & \multicolumn{6}{c}{Score / Success (\%)} \\

        \textbf{Method} & \textbf{Embodied PT.} & \textbf{Generalization} & \textbf{Precision} & \textbf{Long-Horizon} & \textbf{Memory} & \textbf{Open} & \textbf{Average} \\
        \midrule
        GigaWorld-Policy~\cite{ye2026gigaworldpolicy} & Yes & 5.34 / 2.89 & 6.15 / 1.83 & 15.51 / 8.92 & 3.46 / 2.22 & 0.54 / 0.50 & 6.20 / 3.27 \\
        LDA-1B~\cite{lyu2026lda} & Yes & 0.71 / 0.17 & 3.21 / 0.50 & 1.92 / 0.08 & 2.08 / 1.78 & 0.00 / 0.00 & 1.58 / 0.51 \\
        RDT-1B~\cite{liu2024rdt} & Yes & 0.56 / 0.33 & 0.38 / 0.00 & 1.13 / 0.00 & 0.49 / 0.33 & 0.00 / 0.00 & 0.51 / 0.13 \\
        H-RDT~\cite{bi2025hrdt} & Yes & 0.49 / 0.22 & 0.41 / 0.00 & 2.23 / 0.17 & 0.12 / 0.11 & 0.08 / 0.08 & 0.67 / 0.12 \\
        X-WAM~\cite{guo2026xwam} & Yes & \textbf{7.39} / 3.33 & \textbf{6.72} / 1.83 & \textbf{17.47} / 9.08 & \textbf{6.32} / 4.67 & \textbf{0.57} / 0.25 & \textbf{7.69} / 3.83 \\
        \midrule
        HDR-WAM~\cite{qian2026hdr} & No & 4.98 / 3.17 & 5.90 / 2.25 & 9.85 / 4.75 & 6.65 / 4.67 & 0.50 / 0.50 & 5.47 / 3.00 \\
        AHA-WAM~\cite{cai2026ahawam} & No & 5.79 / 3.28 & 5.86 / 2.42 & 8.61 / 2.67 & 2.97 / 2.78 & 0.88 / 0.83 & 4.82 / 2.39 \\
        Fast-WAM~\cite{yuan2026fastwam} & No & 2.34 / 1.11 & 1.96 / 0.00 & 9.14 / 5.17 & 3.55 / 3.44 & 0.42 / 0.42 & 3.48 / 2.03 \\
        ACT~\cite{zhao2023act} & No & 0.69 / 0.56 & 0.85 / 0.00 & 1.73 / 0.92 & 1.65 / 0.13 & 0.00 / 0.00 & 0.98 / 0.32 \\
        \textbf{LAW (Ours)} & \textbf{No} & \textbf{5.93 / 3.39} & \textbf{6.27 / 2.58} & \textbf{11.16 / 6.67} & \textbf{7.46 / 5.44} & \textbf{1.53 / 1.00} & \textbf{6.47 / 3.82} \\
        \bottomrule
    \end{tabular}
    \end{adjustbox}%

\end{table*}

Table~\ref{tab:robodojo-main} reports LAW's performance on RoboDojo across Generalization, Precision, Long-Horizon, Memory, and Open. LAW achieves leading scores and success rates across all five dimensions among the listed methods without additional robot-domain or embodied-interaction pretraining, averaging 6.47 / 3.82\%. On Long-Horizon, LAW reaches 11.16 / 6.67\%, exceeding pretrained LDA-1B, RDT-1B, and H-RDT. On Memory, it achieves 7.46 / 5.44\%, exceeding the pretrained X-WAM (6.32 / 4.67\%). These results support the effectiveness of semantic subtask coordination and hierarchical memory for sustained execution and history-dependent decisions.

\begin{figure*}[t]
    \centering
    \includegraphics[width=\textwidth]{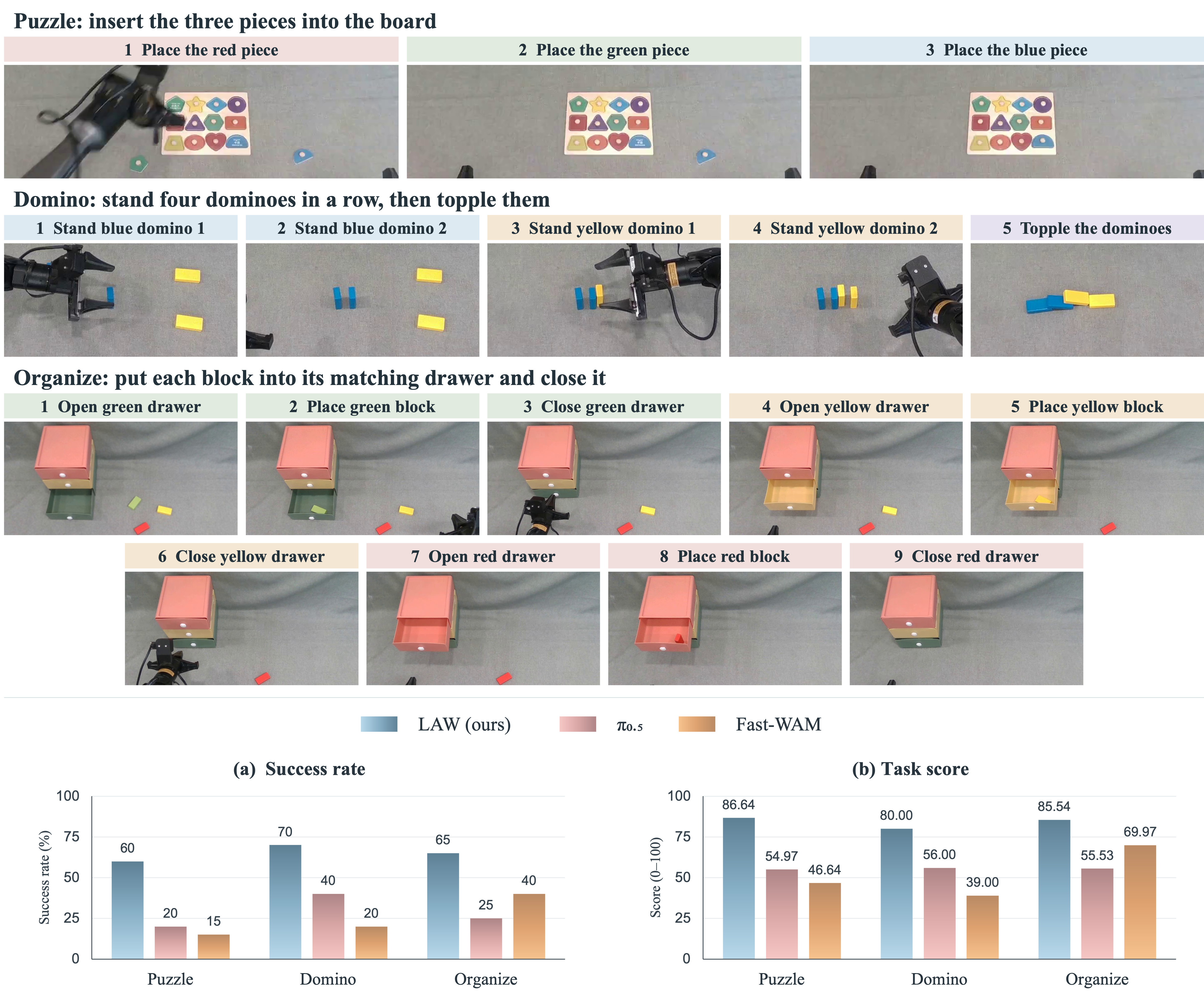}
    \caption{Real-world manipulation results. Representative executions illustrate the semantic subtasks of Puzzle (three), Domino (five), and Organize (nine). (a) Complete-task success rate. (b) Task score, computed from the fraction of completed subtasks and scaled to 0--100. LAW is compared with $\pi_{0.5}$ and Fast-WAM.}
    \label{fig:real-world-results}
\end{figure*}

We further evaluate LAW on three real-world manipulation tasks with distinct sequential requirements: \emph{Puzzle}, which involves inserting three pieces into a board; \emph{Domino}, which requires standing four dominoes in a row and then toppling them; and \emph{Organize}, which requires placing three colored blocks into their matching drawers and closing each drawer. These tasks comprise three, five, and nine semantic subtasks, respectively. We compare LAW with the VLA baseline $\pi_{0.5}$ and the WAM baseline Fast-WAM using complete-task success rate and a task score that assigns equal credit to completed subtasks and normalizes progress to 0--100. Each method is evaluated over 20 trials per task. As shown in Fig.~\ref{fig:real-world-results}, LAW outperforms both baselines on every task under both metrics. Averaged across the three tasks, LAW achieves a success rate of 65\% and a task score of 84.06, compared with 28.33\% and 55.50 for $\pi_{0.5}$, and 25\% and 51.87 for Fast-WAM. On Organize, the task with the longest subtask sequence, LAW achieves 65\% success and a score of 85.54, exceeding the higher-performing of the two baselines, Fast-WAM, by 25 percentage points and 15.57 points, respectively. These results demonstrate improved complete-task execution and subtask completion across real-world manipulation sequences of varying length.

\subsection{Component Ablations}
We progressively add the hybrid-expert architecture, supervision with failure trajectories as negative samples, and hierarchical memory to isolate their effects on VLM planning and state estimation. We evaluate the planner on held-out randomized scenes unseen during training and assess its ability to predict the remaining subtask sequence at different stages of execution from visual observations and history. Planning is measured by normalized edit distance (NED), BLEU, and normalized exact match (EM) of the predicted subtask sequence. 

\begin{table*}[!tb]
    \centering
    \caption{Ablation of VLM experts, failure data, and hierarchical memory.}
    \label{tab:vlm-component-ablation}
    \renewcommand{\arraystretch}{1.25}
    \begin{adjustbox}{max width=\textwidth,max totalheight=0.88\textheight}
    \lawtablefont
\begin{tabular*}{\textwidth}{@{\extracolsep{\fill}}ccc|ccc|ccc@{}}
        \toprule
        \multicolumn{3}{c|}{\textbf{Settings}} & \multicolumn{3}{c|}{\textbf{Planning (\%)}} & \multicolumn{3}{c}{\textbf{State Estimation (\%)}} \\
        \midrule
        \textbf{Hybrid Expert} & \textbf{Failure Data} & \textbf{Memory} & \textbf{NED$\downarrow$} & \textbf{BLEU$\uparrow$} & \textbf{EM$\uparrow$} & \textbf{Accuracy$\uparrow$} & \textbf{Recall$\uparrow$} & \textbf{Weighted F1$\uparrow$} \\
        \midrule
        \multicolumn{3}{c|}{\textbf{Text-only baseline}} & 16.80 & 83.30 & 70.10 & -- & -- & -- \\
        \midrule
        $\times$ & $\times$ & $\times$ & 1.35 & 98.12 & 96.36 & 85.78 & 67.27 & 86.77 \\
        $\surd$ & $\times$ & $\times$ & 0.34 & 99.58 & 99.27 & 98.10 & 95.48 & 98.12 \\
        $\surd$ & $\surd$ & $\times$ & 0.34 & 99.58 & 99.27 & 98.48 & 95.63 & 98.49 \\
        $\bm{\surd}$ & $\bm{\surd}$ & $\bm{\surd}$ & \textbf{0.12} & \textbf{99.77} & \textbf{99.64} & \textbf{99.02} & \textbf{97.13} & \textbf{99.02} \\
        \bottomrule
    \end{tabular*}
    \end{adjustbox}
\end{table*}

\begin{figure*}[t]
    \centering
    \includegraphics[width=0.98\textwidth]{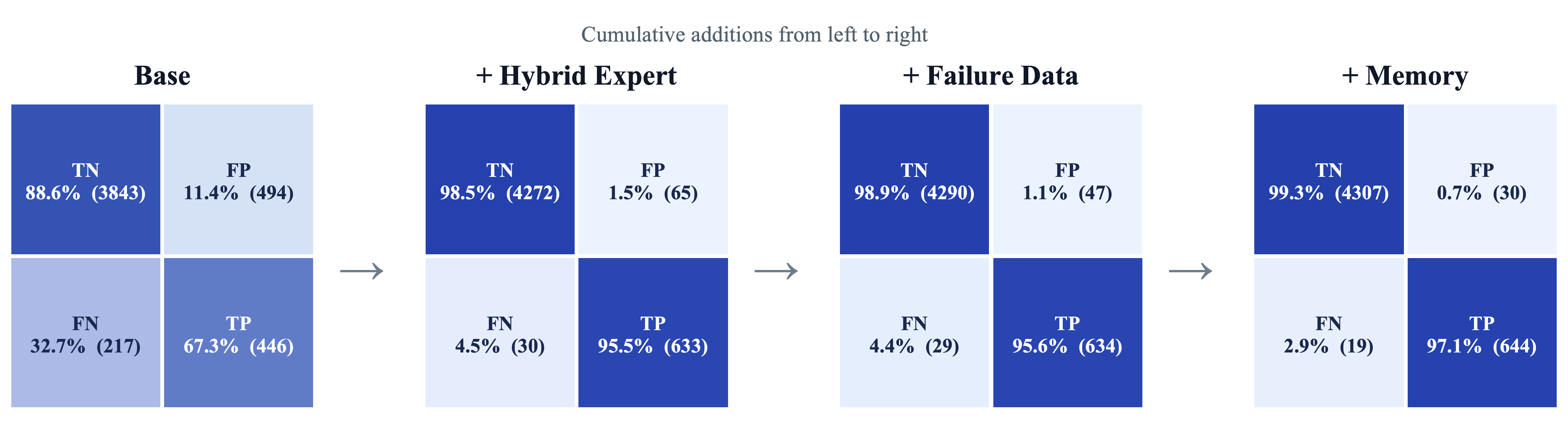}
    \caption{Cumulative ablation of subtask-completion prediction from left to right. Each row-normalized matrix adds one component to the preceding setting; cells report true negatives (TN), false positives (FP), false negatives (FN), and true positives (TP), with sample counts in parentheses.}
    \label{fig:state-confusion-ablation}
\end{figure*}

Table~\ref{tab:vlm-component-ablation} shows that separating control functions with hybrid experts substantially improves both planning and state estimation. Adding failure trajectories as negative samples further reduces completion errors, while hierarchical memory achieves the strongest overall performance. Figure~\ref{fig:state-confusion-ablation} provides a more detailed view of this progression. The false-positive rate decreases from 11.4\% to 0.7\%, and the false-negative rate decreases from 32.7\% to 2.9\%. Completion accuracy increases from 85.78\% to 99.02\%, recall from 67.27\% to 97.13\%, and weighted F1 from 86.77\% to 99.02\%, reducing premature switches and missed completions. The text-only planning baseline receives only the global instruction and completed subtasks, without images, and is evaluated only on planning metrics. It achieves an exact match rate of 70.10\%, compared with 99.64\% for the full planner. This gap indicates that the planner uses current observations for remaining-subtask prediction rather than merely memorizing task templates.

\begin{table}[!tb]
    \centering
    \caption{Ablation of LAW components on RoboTwin.}
    \label{tab:system-component-ablation}
    \setlength{\tabcolsep}{3pt}
    \renewcommand{\arraystretch}{1.25}
    \begin{adjustbox}{max width=\columnwidth,max totalheight=0.88\textheight}
    \lawtablefont
\begin{tabular}{@{}cccc|ccc@{}}
        \toprule
        \textbf{MoTE} & \textbf{Subtask} & \textbf{Memory} & \textbf{Value} & \shortstack{\textbf{Clean}\\\textbf{(\%)}} & \shortstack{\textbf{Rand.}\\\textbf{(\%)}} & \shortstack{\textbf{Avg.}\\\textbf{(\%)}} \\
        \midrule
        $\times$ & $\times$ & $\times$ & $\times$ & 91.88 & 91.78 & 91.83 \\
         $\surd$ & $\times$ & $\times$ & $\times$ & 93.02 & 92.31 & 92.67 \\
        $\times$ & $\surd$ & $\times$ & $\times$ & 92.84 & 92.52 & 92.68 \\
        $\surd$ & $\surd$ & $\times$ & $\times$ & 93.66 & 92.66 & 93.16 \\
        $\surd$ & $\surd$ & $\surd$ & $\times$ & 93.78 & 93.82 & 93.80 \\
        $\bm{\surd}$ & $\bm{\surd}$ & $\bm{\surd}$ & $\bm{\surd}$ & \textbf{94.22} & \textbf{94.20} & \textbf{94.21} \\
        \bottomrule
    \end{tabular}
    \end{adjustbox}
\end{table}

We ablate the components that improve end-to-end task success. Table~\ref{tab:system-component-ablation} includes two independent ``only'' variants and a cumulative sequence. The MoTE-only and subtask-only rows add a single component to Fast-WAM, whereas each subsequent row retains all components enabled above it. MoTE and subtask conditioning provide comparable gains when introduced independently. Combining them raises the average success rate to 93.16\%, hierarchical memory further improves it to 93.80\%, and value-guided action selection achieves the best result of 94.21\%. These gains support complementary benefits from joint video--action modeling, semantic structure, memory, and value guidance.

\section{Analysis}
\label{sec:analysis}

\subsection{WAM Backbone Extensibility}

LAW interacts with the action generator through a common WAM interface, allowing its planning, state estimation, value selection, and memory components to be combined with different WAM backbones. The same value expert is reused across Fast-WAM and OpenWAM-$\alpha$ without backbone-specific adaptation. Despite the difference between recorded training actions and WAM-sampled inference candidates, it improves action selection without additional training on candidates generated by the deployment WAM. Table~\ref{tab:wam-backbone-extensibility} evaluates this extensibility on RoboTwin using the default Fast-WAM backbone and OpenWAM-$\alpha$~\cite{wang2026openwam}. The OpenWAM variant uses the same LAW configuration as the Fast-WAM variant, with only the WAM backbone replaced. LAW improves the average success rate of Fast-WAM from 91.83\% to 94.21\%, and that of OpenWAM-$\alpha$ from 93.60\% to 95.56\%. LAW + OpenWAM-$\alpha$ establishes a new open-source state of the art on RoboTwin with 95.56\% average success. These results demonstrate LAW's extensibility across WAM backbones: its progress-aware control mechanisms improve both action generators, while integration with a stronger pretrained WAM further raises the framework's achievable performance.

\begin{table}[!tb]
    \centering
    \caption{Comparison of WAM backbones on RoboTwin.}
    \label{tab:wam-backbone-extensibility}
    \renewcommand{\arraystretch}{1.22}
    \begin{adjustbox}{max width=\columnwidth,max totalheight=0.88\textheight}
    \lawtablefont
\begin{tabular*}{\columnwidth}{@{\extracolsep{\fill}}l|ccc@{}}
        \toprule
        \textbf{Method} & \textbf{Clean (\%)} & \textbf{Rand. (\%)} & \textbf{Avg. (\%)} \\
        \midrule
        Flex-$\pi$~\cite{yan2026flexpi} & 94.30 & 94.80 & 94.60 \\
        MotuBrain~\cite{motubrainteam2026motubrain}$^{\dagger}$ & 95.80 & 96.08 & 95.94 \\
        \midrule
        Fast-WAM & 91.88 & 91.78 & 91.83 \\
        \textbf{LAW + Fast-WAM} & \textbf{94.22} & \textbf{94.20} & \textbf{94.21} \\
        \midrule
        OpenWAM-$\alpha$ & 93.74 & 93.46 & 93.60 \\
        \textbf{LAW + OpenWAM-$\alpha$} & \textbf{95.60} & \textbf{95.52} & \textbf{95.56} \\
        \bottomrule
    \end{tabular*}
    \end{adjustbox}
\par\vspace{2pt}
    {\footnotesize\raggedright $\dagger$: closed-source methods}
\end{table}

\subsection{Effect of VLM Choice}
\label{sec:vlm-choice}

We compare VLMs for remaining-subtask planning and completion assessment while keeping the WAM backbone and other execution settings fixed. Value-guided action fusion is disabled in all three settings. Table~\ref{tab:vlm-comparison} compares the default 2B hybrid-expert VLM with variants using Embodied-R1.5~\cite{yuan2026embodied} and GPT-5.5 High. The default VLM achieves 93.80\% average success on RoboTwin, compared with 88.8\% and 88.7\%, respectively, corresponding to gains of 5.00 and 5.10 percentage points. It also performs better in both evaluation settings.

With value-guided action fusion disabled, our 2B hybrid-expert VLM achieves higher task success than Embodied-R1.5 and GPT-5.5 High in both evaluation settings. These results highlight the effectiveness of specialized planning and completion assessment for closed-loop manipulation. The ablation in Table~\ref{tab:vlm-component-ablation} further shows that separating planning and completion assessment into specialized experts improves both capabilities. 

\begin{table}[!tb]
    \centering
    
    \caption{Effect of VLM choice on RoboTwin.}
    \label{tab:vlm-comparison}
    \renewcommand{\arraystretch}{1.25}
    \begin{adjustbox}{max width=\columnwidth,max totalheight=0.88\textheight}
    \lawtablefont
\begin{tabular*}{\columnwidth}{@{\extracolsep{\fill}}l|ccc@{}}
        \toprule
        \textbf{VLM} & \textbf{Clean (\%)} & \textbf{Rand. (\%)} & \textbf{Avg. (\%)} \\
        \midrule
        Embodied R1.5 & 88.6 & 89.0 & 88.8 \\
        GPT-5.5 High & 89.7 & 87.7 & 88.7 \\
        \textbf{LAW VLM (Ours)} & \textbf{93.78} & \textbf{93.82} & \textbf{93.80} \\
        \bottomrule
    \end{tabular*}
    \end{adjustbox}
\end{table}

\begin{samepage}
\subsection{MoE Placement and Shared Experts}

We investigate whether MoTE's gains reflect model capacity or expert placement in joint video--action modeling. Table~\ref{tab:moe-analysis} compares MoE layers inserted at conventional attention and FFN locations with MoTE applied after MoT mixed attention. Attn-MoE, FFN-MoE, and MoTE are matched in total and active parameter counts. In MoTE without shared experts, each modality receives one additional private expert in place of the shared expert, resulting in a larger total parameter count but the same active parameter count as MoTE. The no-MoE baseline is excluded from parameter matching. Attn-MoE and FFN-MoE add expert capacity but remain close to the no-MoE baseline, indicating limited benefits from additional parameters alone.
\end{samepage}

\begin{table}[!tb]
    \centering
    
    \caption{Effect of MoE placement and shared experts on RoboTwin.}
    \label{tab:moe-analysis}
    \renewcommand{\arraystretch}{1.22}
    \begin{adjustbox}{max width=\columnwidth,max totalheight=0.88\textheight}
    \lawtablefont
\begin{tabular*}{\columnwidth}{@{\extracolsep{\fill}}l|ccc@{}}
        \toprule
        \textbf{Setting} & \textbf{Clean (\%)} & \textbf{Rand. (\%)} & \textbf{Avg. (\%)} \\
        \midrule
        W/o MoE & 91.88 & 91.78 & 91.83 \\
        Attn-MoE & 91.77 & 91.92 & 91.85 \\
        FFN-MoE & 92.03 & 91.85 & 91.94 \\
        MoTE w/o shared & 91.98 & 92.07 & 91.93 \\
        \textbf{MoTE} & \textbf{93.02} & \textbf{92.31} & \textbf{92.67} \\
        \bottomrule
    \end{tabular*}
    \end{adjustbox}
\end{table}

\begin{figure}[!tb]
    \centering
    \includegraphics[width=\columnwidth]{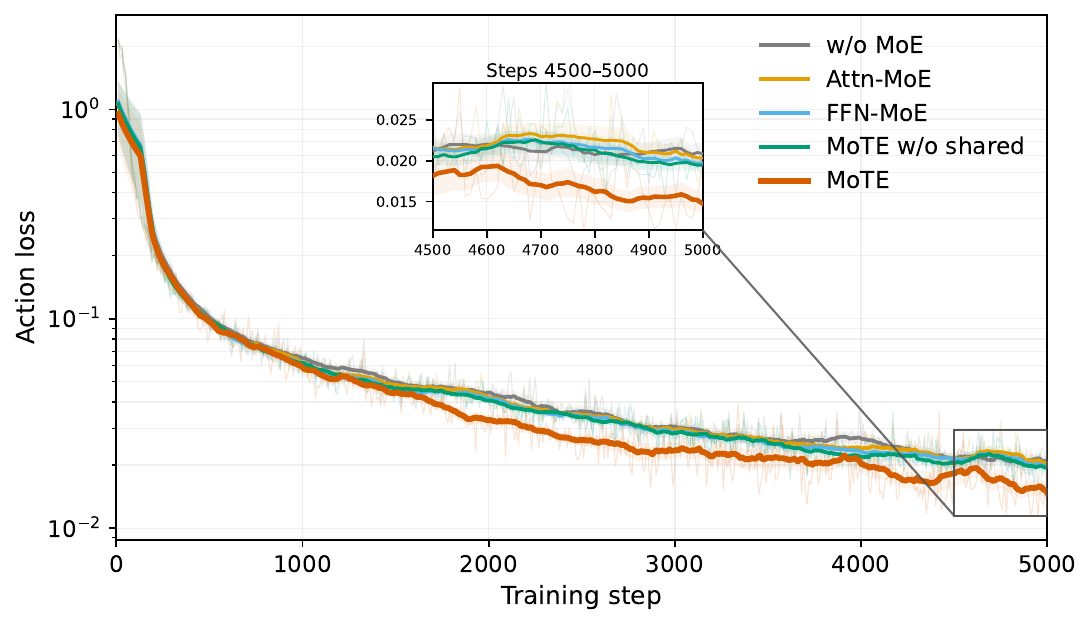}
    \caption{Action-loss curves for different MoE configurations.}
    \label{fig:wam-moe-loss}
\end{figure}
The placement of MoTE is critical because it transforms the representation immediately after video and action tokens exchange information through mixed attention. Experts placed in generic attention or FFN positions mainly increase per-branch capacity, whereas MoTE operates on representations that already contain cross-modal information. Figure~\ref{fig:wam-moe-loss} illustrates the observed loss trends: after the common early descent, full MoTE maintains the lowest action loss throughout the later stage. Attn-MoE and FFN-MoE remain close to the no-MoE baseline, while removing the shared expert also yields a higher action loss than full MoTE. This behavior is consistent with the success-rate comparison in Table~\ref{tab:moe-analysis}. The comparison with MoTE without a shared expert further supports this interpretation. Adding the shared expert improves the average success rate from 91.93\% to 92.67\%. The modality-specific experts preserve the distinct statistics of video and action tokens, while the shared expert captures transformations common to both modalities. Lower loss and higher success rates support joint video--action modeling beyond added capacity.

\subsection{Subtask Granularity}

We compare task-level control, low-level subtasks, and semantic subtasks to examine how subtask granularity affects long-horizon manipulation. The model without subtasks directly conditions control on the overall instruction. Low-level subtasks introduce finer stage partitions, whereas semantic-level subtasks align each stage with a semantic objective shared by the planner, state expert, and memory system. Table~\ref{tab:subtask-granularity-tasks} compares ten tasks.

\begin{table*}[!tb]
    \centering
    \caption{Subtask-granularity ablation on ten RoboTwin tasks.}
    \label{tab:subtask-granularity-tasks}
    \renewcommand{\arraystretch}{1.18}
    \setlength{\tabcolsep}{3pt}
    \begin{adjustbox}{max width=\textwidth,max totalheight=0.88\textheight}
    \lawtablefont
\begin{tabular}{@{}l|ccc|ccc|ccc@{}}
        \toprule
        \textbf{Task} & \multicolumn{3}{c|}{\textbf{W/o Subtask}}
        & \multicolumn{3}{c|}{\textbf{Low-Level Subtasks}}
        & \multicolumn{3}{c}{\textbf{Semantic-Level Subtasks}} \\
        \cline{2-10}
        & \textbf{Clean (\%)} & \textbf{Rand. (\%)} & \textbf{Avg. (\%)} & \textbf{Clean (\%)} & \textbf{Rand. (\%)} & \textbf{Avg. (\%)} & \textbf{Clean (\%)} & \textbf{Rand. (\%)} & \textbf{Avg. (\%)} \\
        \midrule
        Place Can Basket       & 71 & 69 & 70.0 & 54 & 60 & 57.0 & \textbf{85} & \textbf{88} & \textbf{86.5} \\
        Handover Block         & 95 & 81 & 88.0 & 69 & 82 & 75.5 & \textbf{100} & \textbf{96} & \textbf{98.0} \\
        Rotate QRcode          & \textbf{93} & 89 & 91.0 & 87 & 91 & 89.0 & \textbf{93} & \textbf{97} & \textbf{95.0} \\
        Open Laptop            & 98 & \textbf{100} & 99.0 & 99 & 95 & 97.0 & \textbf{100} & 99 & \textbf{99.5} \\
        Stack Blocks Two       & \textbf{100} & \textbf{100} & \textbf{100.0} & 91 & 98 & 94.5 & \textbf{100} & \textbf{100} & \textbf{100.0} \\
        Place Burger Fries     & 96 & \textbf{99} & \textbf{97.5} & \textbf{99} & 96 & \textbf{97.5} & 97 & 98 & \textbf{97.5} \\
        Put Bottles Dustbin    & \textbf{95} & 90 & 92.5 & 92 & \textbf{94} & \textbf{93.0} & 91 & \textbf{94} & 92.5 \\
        Place Container Plate  & 96 & \textbf{100} & 98.0 & \textbf{100} & 97 & \textbf{98.5} & 98 & 98 & 98.0 \\
        Place Object Basket    & 89 & 88 & 88.5 & 89 & 90 & 89.5 & \textbf{93} & \textbf{92} & \textbf{92.5} \\
        Move Pillbottle Pad    & \textbf{100} & \textbf{99} & \textbf{99.5} & \textbf{100} & \textbf{99} & \textbf{99.5} & 98 & \textbf{99} & 98.5 \\
        \midrule
        \textbf{10-task Average}
        & 93.3 & 91.5 & 92.4
        & 88.0 & 90.2 & 89.1
        & \textbf{95.5} & \textbf{96.1} & \textbf{95.8} \\
        \bottomrule
    \end{tabular}
    \end{adjustbox}
    
\end{table*}

Semantic-level subtasks achieve the highest ten-task average, improving from 92.4\% without subtasks and 89.1\% with low-level subtasks to 95.8\%. The largest gains appear on tasks such as Handover Block, Place Can Basket, and Place Object Basket, where progress depends on recognizing distinct manipulation stages. Finer partitions can fragment a coherent semantic objective into short transitions, making completion decisions and historical context less stable. Semantic stages instead provide consistent units for planning, asynchronous state assessment, and hierarchical memory. The semantic decomposition does not improve every saturated task, but its aggregate advantage indicates that the abstraction level of the subtask, rather than decomposition alone, is important for long-horizon control.

\subsection{Subtask-Local Value Learning Ablation}
\label{sec:value-learning-ablation}

We compare action selection without value guidance, value experts trained with episode-level supervision, and value experts trained with subtask-local supervision on five selected multi-subtask RoboTwin tasks. Table~\ref{tab:value-supervision} reports the resulting success rates under clean and randomized settings.

\begin{table*}[!tb]
    \centering
    \caption{Value-supervision ablation on five multi-subtask RoboTwin tasks.}
    \label{tab:value-supervision}
    \renewcommand{\arraystretch}{1.18}
    \setlength{\tabcolsep}{3pt}
    \begin{adjustbox}{max width=\textwidth,max totalheight=0.88\textheight}
    \lawtablefont
    \begin{tabular}{@{}l|ccc|ccc|ccc@{}}
        \toprule
        \textbf{Task}
        & \multicolumn{3}{c|}{\textbf{W/o Value Guidance}}
        & \multicolumn{3}{c|}{\textbf{Episode-Level Supervision}}
        & \multicolumn{3}{c}{\textbf{Subtask-Level Supervision}} \\
        \cline{2-10}
        & \textbf{Clean (\%)} & \textbf{Rand. (\%)} & \textbf{Avg. (\%)}
        & \textbf{Clean (\%)} & \textbf{Rand. (\%)} & \textbf{Avg. (\%)}
        & \textbf{Clean (\%)} & \textbf{Rand. (\%)} & \textbf{Avg. (\%)} \\
        \midrule
        Blocks Ranking Size
        & 68 & 72 & 70.0
        & 72 & 71 & 71.5
        & \textbf{79} & \textbf{75} & \textbf{77.0} \\
        Put Object Cabinet
        & 89 & 87 & 88.0
        & 90 & 85 & 87.5
        & \textbf{91} & \textbf{89} & \textbf{90.0} \\
        Place Burger Fries
        & 98 & 97 & 97.5
        & 97 & \textbf{99} & 98.0
        & \textbf{99} & \textbf{99} & \textbf{99.0} \\
        Stack Bowls Three
        & 86 & \textbf{88} & 87.0
        & 85 & 81 & 83.0
        & \textbf{89} & 87 & \textbf{88.0} \\
        Place Cans Plasticbox
        & \textbf{100} & 97 & 98.5
        & \textbf{100} & 99 & 99.5
        & \textbf{100} & \textbf{100} & \textbf{100.0} \\
        \midrule
        \textbf{5-task Average}
        & 88.20 & 88.20 & 88.20
        & 88.80 & 87.00 & 87.90
        & \textbf{91.60} & \textbf{90.00} & \textbf{90.80} \\
        \bottomrule
    \end{tabular}
    \end{adjustbox}
\end{table*}

On these five tasks, subtask-local supervision achieves 90.80\% average success, compared with 88.20\% without value guidance and 87.90\% with episode-level supervision, improving by 2.60 and 2.90 percentage points, respectively. Its aggregate gains hold in both clean and randomized settings. Compared with episode-level supervision, it improves Blocks Ranking Size from 71.5\% to 77.0\%, Put Object Cabinet from 87.5\% to 90.0\%, and Stack Bowls Three from 83.0\% to 88.0\%. These results support assigning credit at semantic subtask boundaries: completed stages of failed trajectories retain positive supervision, allowing the value expert to evaluate candidate actions against the current objective rather than relying solely on the final episode outcome.

\subsection{Memory Component Ablation}

We cumulatively ablate the three components of hierarchical memory on the RoboDojo Memory dimension. Short-term memory retains the recent observation window, long-term memory adds compressed older observations, and subtask-boundary memory further preserves full-resolution observations at semantic transitions.

\begin{table}[!tb]
    \centering
    
    \caption{Hierarchical-memory ablation on RoboDojo.}
    \label{tab:memory-component-ablation}
    \renewcommand{\arraystretch}{1.22}
    \begin{adjustbox}{max width=\columnwidth,max totalheight=0.88\textheight}
    \lawtablefont
\begin{tabular*}{\columnwidth}{@{\extracolsep{\fill}}l|cc@{}}
        \toprule
        \textbf{Setting} & \textbf{Score} & \textbf{Success (\%)} \\
        \midrule
        W/o memory & 4.06 & 3.44 \\
        + Short-term memory & 5.57 & 4.00 \\
        + Long-term memory & 6.48 & 4.89 \\
        \textbf{+ Subtask-boundary memory} & \textbf{7.46} & \textbf{5.44} \\
        \bottomrule
    \end{tabular*}
    \end{adjustbox}
\end{table}

Table~\ref{tab:memory-component-ablation} shows consistent gains from adding each memory component. Short-term memory raises the score from 4.06 to 5.57 and the success rate from 3.44\% to 4.00\%, indicating the value of recent observations for history-dependent decisions. Adding compressed long-term memory further improves these metrics to 6.48 and 4.89\%, suggesting that older context complements the recent observation window. Preserving full-resolution subtask-boundary observations further improves the score and success rate to 7.46 and 5.44\%. These cumulative improvements support retaining both recent visual details and broader historical context, while highlighting the additional benefit of preserving observations at semantic transitions.

\section{Conclusion}

This paper presented LAW, a closed-loop VLM--WAM framework that unifies language reasoning, action generation, and world modeling through semantic subtask progress. The LAW agent coordinates specialized planning, completion assessment, and value-guided action fusion at complementary frequencies, connecting execution feedback to continued control and subtask transitions. MoTE strengthens joint video--action modeling through shared and modality-specific experts, while dual-stream hierarchical memory preserves recent context and semantic boundary observations. Subtask-local value learning further retains useful supervision from completed stages of failed trajectories. Integrating OpenWAM-$\alpha$ raises RoboTwin success to 95.56\%, establishing a new open-source state of the art and demonstrating LAW's extensibility to stronger pretrained WAMs for further performance gains. These results highlight semantic subtask progress as an effective interface for agentic physical intelligence. Future work will investigate transfer across robot embodiments and adaptive subtask decomposition for longer and more diverse tasks.

\bibliographystyle{ieeenat_fullname}
\bibliography{iclr2027_conference}

@inproceedings{hu2021lora,
  title = {{LoRA}: {L}ow-Rank Adaptation of Large Language Models},
  author = {Hu, Edward J. and Shen, Yelong and Wallis, Phillip and Allen-Zhu, Zeyuan and Li, Yuanzhi and Wang, Shean and Wang, Lu and Chen, Weizhu},
  booktitle = {International Conference on Learning Representations},
  year = {2022},
  url = {https://openreview.net/forum?id=nZeVKeeFYf9},
}

@inproceedings{farebrother2024stop,
  title = {Stop Regressing: {T}raining Value Functions via Classification for Scalable Deep {RL}},
  author = {Farebrother, Jesse and Orbay, Jordi and Vuong, Quan and Ali Taiga, Adrien and Chebotar, Yevgen and Xiao, Ted and Irpan, Alex and Levine, Sergey and Castro, Pablo Samuel and Faust, Aleksandra and others},
  booktitle = {Proceedings of the 41st International Conference on Machine Learning},
  series = {Proceedings of Machine Learning Research},
  volume = {235},
  pages = {13049--13071},
  year = {2024},
  publisher = {PMLR},
  url = {https://proceedings.mlr.press/v235/farebrother24a.html},
}

@article{markley2007averaging,
  title = {Averaging Quaternions},
  author = {Markley, F. Landis and Cheng, Yang and Crassidis, John L. and Oshman, Yaakov},
  journal = {Journal of Guidance, Control, and Dynamics},
  volume = {30},
  number = {4},
  pages = {1193--1197},
  year = {2007},
  doi = {10.2514/1.28949},
  url = {https://doi.org/10.2514/1.28949},
}

@misc{ebert2018visual,
  title = {Visual Foresight: {M}odel-Based Deep Reinforcement Learning for Vision-Based Robotic Control},
  author = {Frederik Ebert and Chelsea Finn and Sudeep Dasari and Annie Xie and Alex Lee and Sergey Levine},
  year = {2018},
  note = {arXiv:1812.00568},
  url = {https://arxiv.org/abs/1812.00568},
}

@inproceedings{hafner2020dream,
  title = {Dream to Control: {L}earning Behaviors by Latent Imagination},
  author = {Hafner, Danijar and Lillicrap, Timothy and Ba, Jimmy and Norouzi, Mohammad},
  booktitle = {International Conference on Learning Representations},
  year = {2020},
  url = {https://openreview.net/forum?id=S1lOTC4tDS},
}

@inproceedings{bar2025nwm,
  title = {Navigation World Models},
  author = {Bar, Amir and Zhou, Gaoyue and Tran, Danny and Darrell, Trevor and LeCun, Yann},
  booktitle = {2025 IEEE/CVF Conference on Computer Vision and Pattern Recognition (CVPR)},
  pages = {15791--15801},
  year = {2025},
  doi = {10.1109/CVPR52734.2025.01472},
  url = {https://openaccess.thecvf.com/content/CVPR2025/html/Bar_Navigation_World_Models_CVPR_2025_paper.html},
}

@misc{assran2025vjepa2,
  title = {{V-JEPA 2}: {S}elf-Supervised Video Models Enable Understanding, Prediction and Planning},
  author = {Assran, Mido and Bardes, Adrien and Fan, David and Garrido, Quentin and Howes, Russell and Komeili, Mojtaba and Muckley, Matthew and Rizvi, Ammar and Roberts, Claire and Sinha, Koustuv and Zholus, Artem and Arnaud, Sergio and Gejji, Abha and Martin, Ada and Hogan, Francois Robert and Dugas, Daniel and Bojanowski, Piotr and Khalidov, Vasil and Labatut, Patrick and Massa, Francisco and Szafraniec, Marc and Krishnakumar, Kapil and Li, Yong and Ma, Xiaodong and Chandar, Sarath and Meier, Franziska and LeCun, Yann and Rabbat, Michael and Ballas, Nicolas},
  year = {2025},
  note = {arXiv:2506.09985},
  url = {https://arxiv.org/abs/2506.09985},
}

@inproceedings{jang2025dreamgen,
  title = {{DreamGen}: {U}nlocking Generalization in Robot Learning through Video World Models},
  author = {Jang, Joel and Ye, Seonghyeon and Lin, Zongyu and Xiang, Jiannan and Bjorck, Johan and Fang, Yu and Hu, Fengyuan and Huang, Spencer and Kundalia, Kaushil and Lin, Yen-Chen and Magne, Lo\"{i}c and Mandlekar, Ajay and Narayan, Avnish and Tan, You Liang and Wang, Guanzhi and Wang, Jing and Wang, Qi and Xu, Yinzhen and Zeng, Xiaohui and Zheng, Kaiyuan and Zheng, Ruijie and Liu, Ming-Yu and Zettlemoyer, Luke and Fox, Dieter and Kautz, Jan and Reed, Scott and Zhu, Yuke and Fan, Linxi},
  booktitle = {Proceedings of The 9th Conference on Robot Learning},
  series = {Proceedings of Machine Learning Research},
  volume = {305},
  pages = {5170--5194},
  year = {2025},
  publisher = {PMLR},
  url = {https://proceedings.mlr.press/v305/jang25a.html},
}

@misc{liang2025videopolicy,
  title = {Video Generators are Robot Policies},
  author = {Junbang Liang and Pavel Tokmakov and Ruoshi Liu and Sruthi Sudhakar and Paarth Shah and Rares Ambrus and Carl Vondrick},
  year = {2025},
  note = {arXiv:2508.00795},
  url = {https://arxiv.org/abs/2508.00795},
}

@misc{liao2026genie,
  title = {{Genie Envisioner}: {A} Unified World Foundation Platform for Robotic Manipulation},
  author = {Liao, Yue and Zhou, Pengfei and Huang, Siyuan and Yang, Donglin and Chen, Shengcong and Jiang, Yuxin and Hu, Yue and Cai, Jingbin and Liu, Si and Luo, Jianlan and Chen, Liliang and Yan, Shuicheng and Yao, Maoqing and Ren, Guanghui},
  year = {2025},
  note = {arXiv:2508.05635},
  url = {https://arxiv.org/abs/2508.05635v3},
}

@misc{zhang2026lingbotva,
  title = {Causal World Modeling for Robot Control},
  author = {Lin Li and others},
  year = {2026},
  note = {arXiv:2601.21998},
  url = {https://arxiv.org/abs/2601.21998},
}

@misc{zhang2026lingbotva2,
  title = {Native Video-Action Pretraining for Generalizable Robot Control},
  author = {Zhang, Qihang and others},
  year = {2026},
  note = {arXiv:2607.08639},
  url = {https://arxiv.org/abs/2607.08639},
}

@misc{yuan2026fastwam,
  title = {{Fast-WAM}: {D}o World Action Models Need Test-time Future Imagination?},
  author = {Yuan, Tianyuan and Dong, Zibin and Liu, Yicheng and Zhao, Hang},
  year = {2026},
  note = {arXiv:2603.16666},
  url = {https://arxiv.org/abs/2603.16666},
}

@inproceedings{zhou2024dinowm,
  title = {{DINO}-{WM}: {W}orld Models on Pre-trained Visual Features enable Zero-shot Planning},
  author = {Zhou, Gaoyue and Pan, Hengkai and LeCun, Yann and Pinto, Lerrel},
  booktitle = {Proceedings of the 42nd International Conference on Machine Learning},
  series = {Proceedings of Machine Learning Research},
  volume = {267},
  pages = {79115--79135},
  year = {2025},
  publisher = {PMLR},
  url = {https://proceedings.mlr.press/v267/zhou25t.html},
}

@inproceedings{zhu2025uwm,
  title = {Unified World Models: {C}oupling Video and Action Diffusion for Pretraining on Large Robotic Datasets},
  author = {Zhu, Chuning and Yu, Raymond and Feng, Siyuan and Burchfiel, Benjamin and Shah, Paarth and Gupta, Abhishek},
  booktitle = {Proceedings of Robotics: Science and Systems},
  year = {2025},
  doi = {10.15607/RSS.2025.XXI.015},
  url = {https://www.roboticsproceedings.org/rss21/p015.html},
}

@inproceedings{zheng2025flare,
  title = {{FLARE}: {R}obot Learning with Implicit World Modeling},
  author = {Zheng, Ruijie and Wang, Jing and Reed, Scott and Bjorck, Johan and Fang, Yu and Hu, Fengyuan and Jang, Joel and Kundalia, Kaushil and Lin, Zongyu and Magne, Lo\"{i}c and Narayan, Avnish and Tan, You Liang and Wang, Guanzhi and Wang, Qi and Xiang, Jiannan and Xu, Yinzhen and Ye, Seonghyeon and Kautz, Jan and Huang, Furong and Zhu, Yuke and Fan, Linxi},
  booktitle = {Proceedings of The 9th Conference on Robot Learning},
  series = {Proceedings of Machine Learning Research},
  volume = {305},
  pages = {3952--3971},
  year = {2025},
  publisher = {PMLR},
  url = {https://proceedings.mlr.press/v305/zheng25a.html},
}

@inproceedings{ahn2022saycan,
  title = {Do As {I} Can, Not As {I} Say: {G}rounding Language in Robotic Affordances},
  author = {Ichter, Brian and Brohan, Anthony and Chebotar, Yevgen and Finn, Chelsea and Hausman, Karol and Herzog, Alexander and Ho, Daniel and Ibarz, Julian and Irpan, Alex and Jang, Eric and Julian, Ryan and Kalashnikov, Dmitry and Levine, Sergey and Lu, Yao and Parada, Carolina and Rao, Kanishka and Sermanet, Pierre and Toshev, Alexander T and Vanhoucke, Vincent and Xia, Fei and Xiao, Ted and Xu, Peng and Yan, Mengyuan and Brown, Noah and Ahn, Michael and Cortes, Omar and Sievers, Nicolas and Tan, Clayton and Xu, Sichun and Reyes, Diego and Rettinghouse, Jarek and Quiambao, Jornell and Pastor, Peter and Luu, Linda and Lee, Kuang-Huei and Kuang, Yuheng and Jesmonth, Sally and Joshi, Nikhil J. and Jeffrey, Kyle and Ruano, Rosario Jauregui and Hsu, Jasmine and Gopalakrishnan, Keerthana and David, Byron and Zeng, Andy and Fu, Chuyuan Kelly},
  booktitle = {Proceedings of The 6th Conference on Robot Learning},
  series = {Proceedings of Machine Learning Research},
  volume = {205},
  pages = {287--318},
  year = {2023},
  publisher = {PMLR},
  url = {https://proceedings.mlr.press/v205/ichter23a.html},
}

@inproceedings{driess2023palme,
  title = {{PaLM-E}: {A}n Embodied Multimodal Language Model},
  author = {Driess, Danny and Xia, Fei and Sajjadi, Mehdi S. M. and Lynch, Corey and Chowdhery, Aakanksha and Ichter, Brian and Wahid, Ayzaan and Tompson, Jonathan and Vuong, Quan and Yu, Tianhe and Huang, Wenlong and Chebotar, Yevgen and Sermanet, Pierre and Duckworth, Daniel and Levine, Sergey and Vanhoucke, Vincent and Hausman, Karol and Toussaint, Marc and Greff, Klaus and Zeng, Andy and Mordatch, Igor and Florence, Pete},
  booktitle = {Proceedings of the 40th International Conference on Machine Learning},
  series = {Proceedings of Machine Learning Research},
  volume = {202},
  pages = {8469--8488},
  year = {2023},
  publisher = {PMLR},
  url = {https://proceedings.mlr.press/v202/driess23a.html},
}

@inproceedings{openx2023rtx,
  title = {{Open X-Embodiment}: {R}obotic Learning Datasets and {RT-X} Models},
  author = {O'Neill, Abby and Rehman, Abdul and Maddukuri, Abhiram and Gupta, Abhishek and Padalkar, Abhishek and Lee, Abraham and Pooley, Acorn and Gupta, Agrim and Mandlekar, Ajay and Jain, Ajinkya and Tung, Albert and Bewley, Alex and others},
  booktitle = {2024 IEEE International Conference on Robotics and Automation (ICRA)},
  pages = {6892--6903},
  year = {2024},
  doi = {10.1109/ICRA57147.2024.10611477},
  url = {https://doi.org/10.1109/ICRA57147.2024.10611477},
}

@inproceedings{li2024roboflamingo,
  title = {Vision-Language Foundation Models as Effective Robot Imitators},
  author = {Li, Xinghang and Liu, Minghuan and Zhang, Hanbo and Yu, Cunjun and Xu, Jie and Wu, Hongtao and Cheang, Chilam and Jing, Ya and Zhang, Weinan and Liu, Huaping and Li, Hang and Kong, Tao},
  booktitle = {International Conference on Learning Representations},
  year = {2024},
  url = {https://proceedings.iclr.cc/paper_files/paper/2024/hash/71639c317fb0bf398835627b4418693e-Abstract-Conference.html},
}

@inproceedings{octo2024,
  title = {{Octo}: {A}n Open-Source Generalist Robot Policy},
  author = {Dibya Ghosh and Homer Rich Walke and Karl Pertsch and Kevin Black and Oier Mees and Sudeep Dasari and Joey Hejna and Tobias Kreiman and Charles Xu and Jianlan Luo and You Liang Tan and Lawrence Yunliang Chen and Quan Vuong and Ted Xiao and Pannag R Sanketi and Dorsa Sadigh and Chelsea Finn and Sergey Levine},
  booktitle = {Proceedings of Robotics: Science and Systems},
  year = {2024},
  address = {Delft, Netherlands},
  doi = {10.15607/RSS.2024.XX.090},
  url = {https://roboticsproceedings.org/rss20/p090.html},
}

@inproceedings{kim2024openvla,
  title = {{OpenVLA}: {A}n Open-Source Vision-Language-Action Model},
  author = {Kim, Moo Jin and Pertsch, Karl and Karamcheti, Siddharth and Xiao, Ted and Balakrishna, Ashwin and Nair, Suraj and Rafailov, Rafael and Foster, Ethan P and Sanketi, Pannag R and Vuong, Quan and Kollar, Thomas and Burchfiel, Benjamin and Tedrake, Russ and Sadigh, Dorsa and Levine, Sergey and Liang, Percy and Finn, Chelsea},
  booktitle = {Proceedings of The 8th Conference on Robot Learning},
  series = {Proceedings of Machine Learning Research},
  volume = {270},
  pages = {2679--2713},
  year = {2025},
  publisher = {PMLR},
  url = {https://proceedings.mlr.press/v270/kim25c.html},
}

@inproceedings{black2024pizero,
  title = {{$\pi_0$}: {A} Vision-Language-Action Flow Model for General Robot Control},
  author = {Kevin Black and Noah Brown and Danny Driess and Adnan Esmail and Michael Robert Equi and Chelsea Finn and Niccolo Fusai and Lachy Groom and Karol Hausman and Brian Ichter and Szymon Jakubczak and Tim Jones and Liyiming Ke and Sergey Levine and Adrian Li-Bell and Mohith Mothukuri and Suraj Nair and Karl Pertsch and Lucy Xiaoyang Shi and Laura Smith and James Tanner and Quan Vuong and Anna Walling and Haohuan Wang and Ury Zhilinsky},
  booktitle = {Proceedings of Robotics: Science and Systems},
  year = {2025},
  month = {June},
  address = {Los Angeles, CA, USA},
  doi = {10.15607/RSS.2025.XXI.010},
  url = {https://www.roboticsproceedings.org/rss21/p010.html},
}

@inproceedings{huang2022innermonologue,
  title = {Inner Monologue: {E}mbodied Reasoning through Planning with Language Models},
  author = {Huang, Wenlong and Xia, Fei and Xiao, Ted and Chan, Harris and Liang, Jacky and Florence, Pete and Zeng, Andy and Tompson, Jonathan and Mordatch, Igor and Chebotar, Yevgen and Sermanet, Pierre and Jackson, Tomas and Brown, Noah and Luu, Linda and Levine, Sergey and Hausman, Karol and Ichter, Brian},
  booktitle = {Proceedings of The 6th Conference on Robot Learning},
  series = {Proceedings of Machine Learning Research},
  volume = {205},
  pages = {1769--1782},
  year = {2023},
  publisher = {PMLR},
  url = {https://proceedings.mlr.press/v205/huang23c.html},
}

@inproceedings{belkhale2024rth,
  title = {{RT-H}: {A}ction Hierarchies using Language},
  author = {Suneel Belkhale and Tianli Ding and Ted Xiao and Pierre Sermanet and Quan Vuong and Jonathan Tompson and Yevgen Chebotar and Debidatta Dwibedi and Dorsa Sadigh},
  booktitle = {Proceedings of Robotics: Science and Systems},
  year = {2024},
  address = {Delft, Netherlands},
  doi = {10.15607/RSS.2024.XX.049},
  url = {https://roboticsproceedings.org/rss20/p049.html},
}

@inproceedings{zhao2025cotvla,
  title = {{CoT-VLA}: {V}isual Chain-of-Thought Reasoning for Vision-Language-Action Models},
  author = {Qingqing Zhao and Yao Lu and Moo Jin Kim and Zipeng Fu and Zhuoyang Zhang and Yecheng Wu and Zhaoshuo Li and Qianli Ma and Song Han and Chelsea Finn and Ankur Handa and Tsung-Yi Lin and Gordon Wetzstein and Ming-Yu Liu and Donglai Xiang},
  booktitle = {Proceedings of the IEEE/CVF Conference on Computer Vision and Pattern Recognition (CVPR)},
  pages = {1702--1713},
  year = {2025},
  month = {June},
  url = {https://openaccess.thecvf.com/content/CVPR2025/html/Zhao_CoT-VLA_Visual_Chain-of-Thought_Reasoning_for_Vision-Language-Action_Models_CVPR_2025_paper.html},
}

@misc{wang2026openwam,
  title = {{OpenWAM}: {A}n Open, Modular Exploration Towards Systematic World-Action Model Pretraining},
  author = {Yuran Wang and Siqiao Huang and Mingleyang Li and Chenhao Zhang and Jiaqi Liang and Weiyang Jin and Yue Chen and Xuemin Chi and Donghao Zhou and Qize Yu and others},
  year = {2026},
  note = {arXiv:2609.07398},
  url = {https://arxiv.org/abs/2609.07398},
}

@misc{chen2026worldsync,
  title = {Do Robotic World Models Really Follow Actions? {D}iagnosing and Aligning Action-Conditioned Generation for Policy Learning},
  author = {Chen, Sixiang and Liu, Jiaming and Wu, Jixian and Guo, Yichen and Wang, Tinghao and Qian, Siyuan and Chen, Hao and Cao, Jiajun and Tang, Jian and Zhang, Shanghang},
  year = {2026},
  note = {arXiv:2608.24885},
  url = {https://arxiv.org/abs/2608.24885},
}

@misc{chen2026mvwam,
  title = {{MV-WAM}: {M}anifold-Aware World Action Model with Value Augmentation},
  author = {Jintao Chen and Peidong Jia and Qingpo Wuwu and Jiaming Liu and Mengfei Du and Chun-Kai Fan and Xiaowei Chi and Hao Chen and Chengyu Bai and Zezhong Qian and Hao Wang and Jiajun Cao and Weishi Mi and Xiaozhu Ju and Jian Tang and Shanghang Zhang},
  year = {2026},
  note = {arXiv:2606.21088},
  url = {https://arxiv.org/abs/2606.21088},
}

@misc{tang2026safedojo,
  title = {{SafeDojo}: {S}afe Reinforcement Learning for {VLA} via Interactive World Model},
  author = {Tang, Kai and Jia, Peidong and Chu, Zhong and Wu, Jixian and Ma, Rui and Cao, Jiajun and Zhao, Fangyuan and Chen, Sixiang and Guo, Yichen and Chi, Xiaowei and Fan, Chun-Kai and Zhang, Kevin and Xu, Jinchang and Yang, Fubing and Mi, Weishi and Ju, Xiaozhu and Tang, Jian and Zhang, Shanghang},
  year = {2026},
  note = {arXiv:2606.20698},
  url = {https://arxiv.org/abs/2606.20698},
}

@misc{wang2026vega,
  title = {{VEGA}: {V}isual Encoder Grounding Alignment for Spatially-Aware Vision-Language-Action Models},
  author = {Wang, Hao and Wei, Xiaobao and He, Jingyang and Bai, Chengyu and Fan, Chun-Kai and Cao, Jiajun and Chen, Jintao and Li, Ying and Rong, Shanyu and Lu, Ming and Ju, Xiaozhu and Tang, Jian and Zhang, Shanghang},
  year = {2026},
  note = {arXiv:2605.10485},
  url = {https://arxiv.org/abs/2605.10485},
}

@misc{li2026wam4d,
  title = {{WAM4D}: {F}ast {4D} World Action Model via Spatial Register Tokens},
  author = {Ying Li and Xiaobao Wei and Jiajun Cao and Hao Wang and Xiaowei Chi and Chengyu Bai and Qianpu Sun and Jiajun Li and Xiaojie Zhang and Peidong Jia and Jian Tang and Sirui Han and Shanghang Zhang},
  year = {2026},
  note = {arXiv:2606.14048},
  url = {https://arxiv.org/abs/2606.14048},
}

@inproceedings{kim2026cosmospolicy,
  title = {{Cosmos Policy}: {F}ine-Tuning Video Models for Visuomotor Control and Planning},
  author = {Kim, Moo Jin and Gao, Yihuai and Lin, Tsung-Yi and Lin, Yen-Chen and Ge, Yunhao and Lam, Grace and Liang, Percy and Song, Shuran and Liu, Ming-Yu and Finn, Chelsea and Gu, Jinwei},
  booktitle = {International Conference on Learning Representations},
  volume = {2026},
  pages = {71531--71552},
  year = {2026},
  url = {https://proceedings.iclr.cc/paper_files/paper/2026/file/748becc400a57c0e31cfe6a2e7951467-Paper-Conference.pdf},
}

@article{lin2025navcot,
  title = {{NavCoT}: {B}oosting {LLM}-Based Vision-and-Language Navigation via Learning Disentangled Reasoning},
  author = {Lin, Bingqian and Nie, Yunshuang and Wei, Ziming and Chen, Jiaqi and Ma, Shikui and Han, Jianhua and Xu, Hang and Chang, Xiaojun and Liang, Xiaodan},
  journal = {IEEE Transactions on Pattern Analysis and Machine Intelligence},
  volume = {47},
  number = {7},
  pages = {5945--5957},
  year = {2025},
  doi = {10.1109/TPAMI.2025.3554559},
  url = {https://doi.org/10.1109/TPAMI.2025.3554559},
}

@article{xu2026memoir,
  title = {Dream to Recall: {I}magination-Guided Experience Retrieval for Memory-Persistent Vision-and-Language Navigation},
  author = {Xu, Yunzhe and Pan, Yiyuan and Liu, Zhe},
  journal = {IEEE Transactions on Pattern Analysis and Machine Intelligence},
  volume = {48},
  number = {8},
  pages = {9035--9049},
  year = {2026},
  doi = {10.1109/TPAMI.2026.3679426},
  url = {https://doi.org/10.1109/TPAMI.2026.3679426},
}

@inproceedings{zitkovich2023rt2,
  title = {{RT-2}: {V}ision-Language-Action Models Transfer Web Knowledge to Robotic Control},
  author = {Zitkovich, Brianna and Yu, Tianhe and Xu, Sichun and Xu, Peng and Xiao, Ted and Xia, Fei and others},
  booktitle = {Proceedings of The 7th Conference on Robot Learning},
  series = {Proceedings of Machine Learning Research},
  volume = {229},
  pages = {2165--2183},
  year = {2023},
  publisher = {PMLR},
  url = {https://proceedings.mlr.press/v229/zitkovich23a.html},
}

@inproceedings{huang2023voxposer,
  title = {{VoxPoser}: {C}omposable {3D} Value Maps for Robotic Manipulation with Language Models},
  author = {Huang, Wenlong and Wang, Chen and Zhang, Ruohan and Li, Yunzhu and Wu, Jiajun and Fei-Fei, Li},
  booktitle = {Proceedings of The 7th Conference on Robot Learning},
  series = {Proceedings of Machine Learning Research},
  volume = {229},
  pages = {540--562},
  year = {2023},
  publisher = {PMLR},
  url = {https://proceedings.mlr.press/v229/huang23b.html},
}

@inproceedings{zheng2025xvla,
  title = {{X-VLA}: {S}oft-Prompted Transformer as Scalable Cross-Embodiment Vision-Language-Action Model},
  author = {Jinliang Zheng and Jianxiong Li and Zhihao Wang and Dongxiu Liu and Xirui Kang and Yuchun Feng and Yinan Zheng and Jiayin Zou and Yilun Chen and Jia Zeng and Tai Wang and Ya-Qin Zhang and Jingjing Liu and Xianyuan Zhan},
  booktitle = {International Conference on Learning Representations},
  year = {2026},
  url = {https://proceedings.iclr.cc/paper_files/paper/2026/hash/630d293833e09e1ecd892a898a20b074-Abstract-Conference.html},
}

@inproceedings{black2025pi05,
  title = {$\pi_{0.5}$: {A} Vision-Language-Action Model with Open-World Generalization},
  author = {Black, Kevin and Brown, Noah and Darpinian, James and Dhabalia, Karan and Driess, Danny and Esmail, Adnan and Equi, Michael Robert and Finn, Chelsea and Fusai, Niccolo and Galliker, Manuel Y. and Ghosh, Dibya and Groom, Lachy and Hausman, Karol and Ichter, Brian and Jakubczak, Szymon and Jones, Tim and Ke, Liyiming and LeBlanc, Devin and Levine, Sergey and Li-Bell, Adrian and Mothukuri, Mohith and Nair, Suraj and Pertsch, Karl and Ren, Allen Z. and Shi, Lucy Xiaoyang and Smith, Laura and Springenberg, Jost Tobias and Stachowicz, Kyle and Tanner, James and Vuong, Quan and Walke, Homer and Walling, Anna and Wang, Haohuan and Yu, Lili and Zhilinsky, Ury},
  booktitle = {Proceedings of The 9th Conference on Robot Learning},
  series = {Proceedings of Machine Learning Research},
  volume = {305},
  pages = {17--40},
  year = {2025},
  publisher = {PMLR},
  url = {https://proceedings.mlr.press/v305/black25a.html},
}

@misc{yuan2026qwenrobotmanip,
  title = {{Qwen-RobotManip} Technical Report: {A}lignment Unlocks Scale for Robotic Manipulation Foundation Models},
  author = {Yuan, Haoqi and Liang, Zhixuan and Chen, Anzhe and Wang, Ye and Li, Haoyang and Lin, Pei and Huang, Yiyang and Lei, Zixing and Zhang, Tong and Zhang, Jiazhao and Zhang, Jie and Fan, Jingyang and Zhou, Gengze and Peng, Qihang and Lv, Chenxu and Chen, Xiaoyue and Yang, An and Huang, Fei and Lin, Junyang and Liu, Dayiheng and Zhou, Jingren and Wu, Chenfei and Chen, Xiong-Hui},
  year = {2026},
  note = {arXiv:2606.17846},
  url = {https://arxiv.org/abs/2606.17846},
}

@inproceedings{bi2026motus,
  title = {{Motus}: {A} Unified Latent Action World Model},
  author = {Bi, Hongzhe and Tan, Hengkai and Xie, Shenghao and Wang, Zeyuan and Huang, Shuhe and Liu, Haitian and Zhao, Ruowen and Feng, Yao and Xiang, Chendong and Rong, Yinze and Zhao, Hongyan and Liu, Hanyu and Su, Zhizhong and Ma, Lei and Su, Hang and Zhu, Jun},
  booktitle = {Proceedings of the IEEE/CVF Conference on Computer Vision and Pattern Recognition (CVPR)},
  pages = {35101--35113},
  year = {2026},
  month = {June},
  url = {https://openaccess.thecvf.com/content/CVPR2026/html/Bi_Motus_A_Unified_Latent_Action_World_Model_CVPR_2026_paper.html},
}

@misc{ye2026gigaworldpolicy,
  title = {{GigaWorld-Policy}: {A}n Efficient Action-Centered World--Action Model},
  author = {Ye, Angen and Wang, Boyuan and Ni, Chaojun and Huang, Guan and Zhao, Guosheng and Li, Hao and Li, Hengtao and Li, Jie and Lv, Jindi and Liu, Jingyu and Cao, Min and Li, Peng and Deng, Qiuping and Mei, Wenjun and Wang, Xiaofeng and Chen, Xinze and Zhou, Xinyu and Wang, Yang and Chang, Yifan and Li, Yifan and Zhou, Yukun and Ye, Yun and Liu, Zhichao and Zhu, Zheng},
  year = {2026},
  note = {arXiv:2603.17240},
  url = {https://arxiv.org/abs/2603.17240},
}

@inproceedings{lyu2026lda,
  title = {{LDA-1B}: {S}caling Latent Dynamics Action Model via Universal Embodied Data Ingestion},
  author = {Lyu, Jiangran and Liu, Kai and Zhang, Xuheng and Liao, Haoran and Feng, Yusen and Zhu, Wenxuan and Shen, Tingrui and Chen, Jiayi and Zhang, Jiazhao and Dong, Yifei and Cui, Wenbo and Qi, Senmao and Wang, Shuo and Zheng, Yixin and Yan, Mi and Shi, Xuesong and Li, Haoran and Zhao, Dongbin and Liu, Ming-Yu and Zhang, Zhizheng and Yi, Li and Wang, Yizhou and Wang, He},
  booktitle = {Proceedings of Robotics: Science and Systems},
  year = {2026},
  address = {Sydney, Australia},
  doi = {10.15607/RSS.2026.XXII.210},
  url = {https://www.roboticsproceedings.org/rss22/p210.html},
}

@inproceedings{liu2024rdt,
  title = {{RDT-1B}: {A} Diffusion Foundation Model for Bimanual Manipulation},
  author = {Liu, Songming and Wu, Lingxuan and Li, Bangguo and Tan, Hengkai and Chen, Huayu and Wang, Zhengyi and Xu, Ke and Su, Hang and Zhu, Jun},
  booktitle = {International Conference on Learning Representations},
  year = {2025},
  url = {https://proceedings.iclr.cc/paper_files/paper/2025/hash/49f80e4d2471ad4f2edf4f5f1ab62339-Abstract-Conference.html},
}

@inproceedings{bi2025hrdt,
  title = {{H-RDT}: {H}uman Manipulation Enhanced Bimanual Robotic Manipulation},
  author = {Bi, Hongzhe and Wu, Lingxuan and Lin, Tianwei and Tan, Hengkai and Su, Zhizhong and Su, Hang and Zhu, Jun},
  booktitle = {Proceedings of the AAAI Conference on Artificial Intelligence},
  volume = {40},
  number = {22},
  pages = {18135--18143},
  year = {2026},
  doi = {10.1609/aaai.v40i22.38875},
  url = {https://ojs.aaai.org/index.php/AAAI/article/view/38875},
}

@misc{guo2026xwam,
  title = {Unified {4D} World Action Modeling from Video Priors with Asynchronous Denoising},
  author = {Guo, Jun and Li, Qiwei and Li, Peiyan and Chen, Zilong and Sun, Nan and Su, Yifei and Wang, Heyun and Zhang, Yuan and Li, Xinghang and Liu, Huaping},
  year = {2026},
  note = {arXiv:2604.26694},
  url = {https://arxiv.org/abs/2604.26694},
}

@misc{qian2026hdr,
  title = {Hierarchical Denoising For Multi-Step Visual Reasoning},
  author = {Zezhong Qian and Xiaowei Chi and Chak-Wing Mak and Tianze Zhou and Ruibin Yuan and Yuhan Rui and Hengzhe Sun and Zhuoqun Wu and Yuming Li and Siyuan Qian and Sirui Han and Shanghang Zhang},
  year = {2026},
  note = {arXiv:2607.15278},
  url = {https://arxiv.org/abs/2607.15278},
}

@misc{cai2026ahawam,
  title = {{AHA-WAM}: {A}synchronous Horizon-Adaptive World-Action Modeling with Observation-Guided Context Routing},
  author = {Cai, Jisong and Ling, Long and Chu, Shiwei and Liu, Zhongshan and Kang, Jiayue and Liang, Zhixuan and Xu, Wenjie and Mao, Yinan and Zhang, Weinan and Yang, Xiaokang and Ying, Ru and Zheng, Ran and Mu, Yao},
  year = {2026},
  note = {arXiv:2606.09811},
  url = {https://arxiv.org/abs/2606.09811},
}

@inproceedings{zhao2023act,
  title = {Learning Fine-Grained Bimanual Manipulation with Low-Cost Hardware},
  author = {Tony Z. Zhao and Vikash Kumar and Sergey Levine and Chelsea Finn},
  booktitle = {Proceedings of Robotics: Science and Systems},
  year = {2023},
  address = {Daegu, Republic of Korea},
  doi = {10.15607/RSS.2023.XIX.016},
  url = {https://roboticsproceedings.org/rss19/p016.html},
}

@inproceedings{liu2023reflect,
  title = {{REFLECT}: {S}ummarizing Robot Experiences for Failure Explanation and Correction},
  author = {Liu, Zeyi and Bahety, Arpit and Song, Shuran},
  booktitle = {Proceedings of The 7th Conference on Robot Learning},
  series = {Proceedings of Machine Learning Research},
  volume = {229},
  pages = {3468--3484},
  year = {2023},
  publisher = {PMLR},
  url = {https://proceedings.mlr.press/v229/liu23g.html},
}

@inproceedings{shi2026memoryvla,
  title = {{MemoryVLA}: {P}erceptual-Cognitive Memory in Vision-Language-Action Models for Robotic Manipulation},
  author = {Shi, Hao and Xie, Bin and Liu, Yingfei and Sun, Lin and Liu, Fengrong and Wang, Tiancai and Zhou, Erjin and Fan, Haoqiang and Zhang, Xiangyu and Huang, Gao},
  booktitle = {International Conference on Learning Representations},
  volume = {2026},
  pages = {18567--18602},
  year = {2026},
  url = {https://proceedings.iclr.cc/paper_files/paper/2026/file/1fa3d6ccbcd15f7285fee666b2bd57be-Paper-Conference.pdf},
}

@misc{torne2026mem,
  title = {{MEM}: {M}ulti-Scale Embodied Memory for Vision Language Action Models},
  author = {Marcel Torne and Karl Pertsch and Homer Walke and Kyle Vedder and Suraj Nair and Brian Ichter and Allen Z. Ren and Haohuan Wang and Jiaming Tang and Kyle Stachowicz and others},
  year = {2026},
  note = {arXiv:2603.03596},
  url = {https://arxiv.org/abs/2603.03596},
}

@misc{chen2025robotwin,
  title = {{RoboTwin 2.0}: {A} Scalable Data Generator and Benchmark with Strong Domain Randomization for Robust Bimanual Robotic Manipulation},
  author = {Tianxing Chen and Zanxin Chen and Baijun Chen and Zijian Cai and Yibin Liu and Zixuan Li and Qiwei Liang and Xianliang Lin and Yiheng Ge and Zhenyu Gu and others},
  year = {2025},
  note = {arXiv:2506.18088},
  url = {https://arxiv.org/abs/2506.18088},
}

@misc{chen2026robodojo,
  title = {{RoboDojo}: {A} Unified Sim-and-Real Benchmark for Comprehensive Evaluation of Generalist Robot Manipulation Policies},
  author = {Chen, Tianxing and Chen, Yue and Li, Zixuan and Tang, Junyuan and Su, Kailun and Lu, Haoran and Wan, Weijie and Chen, Baijun and Liu, Songling and Yan, Haowen and others},
  year = {2026},
  note = {arXiv:2607.04434},
  url = {https://arxiv.org/abs/2607.04434},
}

@misc{yuan2026embodied,
  title = {{Embodied-R1.5}: {E}volving Physical Intelligence via Embodied Foundation Models},
  author = {Yuan, Yifu and Huang, Yaoting and Yao, Xianze and Li, Yutong and Zhang, Shuoheng and Han, Linqi and Li, Pengyi and Sun, Jiangeng and Jia, Wenting and Zhang, Zhao and others},
  year = {2026},
  note = {arXiv:2606.11324},
  url = {https://arxiv.org/abs/2606.11324},
}

@misc{bai2025qwen3,
  title = {{Qwen3-VL} Technical Report},
  author = {Bai, Shuai and Cai, Yuxuan and Chen, Ruizhe and Chen, Keqin and Chen, Xionghui and Cheng, Zesen and Deng, Lianghao and Ding, Wei and Gao, Chang and Ge, Chunjiang and others},
  year = {2025},
  note = {arXiv:2511.21631},
  url = {https://arxiv.org/abs/2511.21631},
}

@misc{yan2026flexpi,
  title = {Flex-{$\pi$}: {A} Multi-Stream World-Action Model with Compute Flexibility},
  author = {Yan, Ge and Liu, Jinghao and Fan, Yuzhi and Cai, Lei and Liao, Minwen and Zhang, Jesse and Fox, Dieter},
  year = {2026},
  note = {arXiv:2608.10860},
  url = {https://arxiv.org/abs/2608.10860},
}

@misc{motubrainteam2026motubrain,
  title = {{Motubrain}: {A}n Advanced World Action Model for Robot Control},
  author = {{Motubrain Team} and Xiang, Chendong and Bao, Fan and Liu, Haitian and Tan, Hengkai and Bi, Hongzhe and Li, James and Liu, Jiabao and Pang, Jingrui and Jing, Kiro and Liu, Louis and Cai, Mengchen and Cui, Rongxu and Zhao, Ruowen and Wang, Runqing and Huang, Shuhe and Feng, Yao and Rong, Yinze and Wang, Zeyuan and Zhu, Jun},
  year = {2026},
  note = {arXiv:2604.27792},
  url = {https://arxiv.org/abs/2604.27792},
}

@article{liang2025mot,
  title = {{Mixture-of-Transformers}: A Sparse and Scalable Architecture for Multi-Modal Foundation Models},
  author = {Liang, Weixin and Yu, Lili and Luo, Liang and Iyer, Srini and Dong, Ning and Zhou, Chunting and Ghosh, Gargi and Lewis, Mike and Yih, Wen-tau and Zettlemoyer, Luke and Lin, Xi Victoria},
  journal = {Transactions on Machine Learning Research},
  year = {2025},
  url = {https://openreview.net/forum?id=Nu6N69i8SB},
  issn = {2835-8856},
}

@misc{wan2025wan22ti2v5b,
  title = {{Wan2.2-TI2V-5B}},
  author = {{Team Wan}},
  year = {2025},
  month = {Jul},
  note = {Official model card and checkpoint, released July 28, 2025},
  url = {https://huggingface.co/Wan-AI/Wan2.2-TI2V-5B},
  howpublished = {Hugging Face},
}

\clearpage
\onecolumn
\appendix
\section{Per-Task RoboTwin Results}
\label{app:robotwin_per_task}
Table~\ref{tab:robotwin-per-task} reports success rates on all 50 RoboTwin tasks for LAW with Fast-WAM and OpenWAM-$\alpha$ backbones, together with the compared baselines. Baseline entries reproduce reported per-task results. Means are computed with equal weight per task and, for the overall score, per evaluation setting. LAW + Fast-WAM achieves 94.22\% and 94.20\% on the clean and randomized settings. LAW + OpenWAM-$\alpha$ achieves 95.60\% and 95.52\%, respectively, yielding an overall average of 95.56\%. The OpenWAM variant uses 100 episodes per task and setting, as in Table~\ref{tab:wam-backbone-extensibility}.

\begin{table}[!tb]
    \centering
    \caption{Per-task results on the 50-task RoboTwin benchmark.}
    \label{tab:robotwin-per-task}
    \renewcommand{\arraystretch}{1.0}
    \setlength{\tabcolsep}{2.2pt}
    \fontsize{8}{9.2}\selectfont
\begin{tabular*}{\textwidth}{@{\extracolsep{\fill}}l*{10}{r}@{}}
        \toprule
        \textbf{Task} & \multicolumn{2}{c}{\textbf{LAW + OpenWAM-$\alpha$}} & \multicolumn{2}{c}{\textbf{LAW + Fast-WAM}} & \multicolumn{2}{c}{\textbf{Fast-WAM}} & \multicolumn{2}{c}{\textbf{LingBot-VA}} & \multicolumn{2}{c}{$\bm{\pi}_{0.5}$} \\
        \cmidrule(lr){2-3}\cmidrule(lr){4-5}\cmidrule(lr){6-7}\cmidrule(lr){8-9}\cmidrule(lr){10-11}
         & \textbf{Clean (\%)} & \textbf{Rand. (\%)} & \textbf{Clean (\%)} & \textbf{Rand. (\%)} & \textbf{Clean (\%)} & \textbf{Rand. (\%)} & \textbf{Clean (\%)} & \textbf{Rand. (\%)} & \textbf{Clean (\%)} & \textbf{Rand. (\%)} \\
        \midrule
        Adjust Bottle & \textbf{100} & \textbf{100} & \textbf{100} & \textbf{100} & \textbf{100} & \textbf{100} & 90 & 94 & \textbf{100} & 99 \\
        Beat Block Hammer & 97 & \textbf{100} & 96 & 99 & \textbf{99} & 97 & 96 & 98 & 96 & 93 \\
        Blocks Ranking RGB & \textbf{100} & 99 & 95 & 93 & \textbf{100} & \textbf{100} & 99 & 98 & 92 & 85 \\
        Blocks Ranking Size & 91 & 91 & 79 & 75 & \textbf{94} & \textbf{98} & \textbf{94} & 96 & 49 & 26 \\
        Click Alarmclock & 99 & 99 & \textbf{100} & 99 & \textbf{100} & \textbf{100} & 99 & \textbf{100} & 98 & 89 \\
        Click Bell & \textbf{100} & \textbf{100} & \textbf{100} & \textbf{100} & \textbf{100} & \textbf{100} & \textbf{100} & \textbf{100} & 99 & 66 \\
        Dump Bin Bigbin & 95 & 91 & 94 & \textbf{97} & \textbf{97} & 96 & 89 & 96 & 92 & \textbf{97} \\
        Grab Roller & \textbf{100} & \textbf{100} & \textbf{100} & \textbf{100} & \textbf{100} & \textbf{100} & \textbf{100} & \textbf{100} & \textbf{100} & \textbf{100} \\
        Handover Block & \textbf{100} & \textbf{97} & \textbf{100} & \textbf{97} & 95 & 81 & 99 & 78 & 66 & 57 \\
        Handover Mic & \textbf{99} & 94 & 97 & 94 & \textbf{99} & \textbf{100} & 94 & 96 & 98 & 97 \\
        Hanging Mug & \textbf{77} & \textbf{78} & 44 & 55 & 58 & 62 & 40 & 28 & 18 & 17 \\
        Lift Pot & \textbf{100} & \textbf{100} & \textbf{100} & \textbf{100} & \textbf{100} & \textbf{100} & \textbf{100} & 99 & 96 & 85 \\
        Move Can Pot & \textbf{100} & \textbf{99} & \textbf{100} & 98 & 90 & 88 & 94 & 97 & 51 & 55 \\
        Move Pillbottle Pad & 99 & 99 & \textbf{100} & \textbf{100} & \textbf{100} & 99 & 99 & 99 & 84 & 61 \\
        Move Playingcard Away & \textbf{100} & \textbf{100} & \textbf{100} & \textbf{100} & \textbf{100} & \textbf{100} & \textbf{100} & 99 & 96 & 84 \\
        Move Stapler Pad & \textbf{98} & \textbf{96} & 86 & 85 & 77 & 64 & 91 & 79 & 56 & 42 \\
        Open Laptop & 99 & 98 & \textbf{100} & 99 & 98 & \textbf{100} & 92 & 94 & 90 & 96 \\
        Open Microwave & \textbf{97} & \textbf{96} & 96 & 95 & 62 & 45 & 82 & 86 & 34 & 77 \\
        Pick Diverse Bottles & \textbf{93} & \textbf{93} & 90 & 90 & 80 & 85 & 89 & 82 & 81 & 71 \\
        Pick Dual Bottles & \textbf{100} & \textbf{100} & \textbf{100} & \textbf{100} & \textbf{100} & 96 & \textbf{100} & 99 & 93 & 63 \\
        Place A2B Left & 94 & \textbf{98} & \textbf{99} & 96 & 95 & 93 & 97 & 93 & 87 & 82 \\
        Place A2B Right & \textbf{99} & \textbf{100} & 98 & 97 & 93 & 99 & 97 & 95 & 87 & 84 \\
        Place Bread Basket & 96 & \textbf{96} & \textbf{99} & 95 & 91 & 93 & 97 & 95 & 77 & 64 \\
        Place Bread Skillet & \textbf{95} & 94 & 94 & \textbf{95} & 90 & 93 & \textbf{95} & 90 & 85 & 66 \\
        Place Burger Fries & 95 & 97 & \textbf{99} & \textbf{99} & 96 & \textbf{99} & 97 & 95 & 94 & 87 \\
        Place Can Basket & \textbf{93} & \textbf{86} & 79 & 84 & 71 & 69 & 81 & 84 & 62 & 62 \\
        Place Cans Plasticbox & \textbf{100} & 99 & \textbf{100} & \textbf{100} & 99 & 96 & \textbf{100} & 99 & 94 & 84 \\
        Place Container Plate & 96 & 99 & 96 & 98 & 96 & \textbf{100} & \textbf{99} & 97 & \textbf{99} & 95 \\
        Place Dual Shoes & \textbf{99} & \textbf{98} & 95 & 96 & 94 & 88 & 94 & 89 & 75 & 75 \\
        Place Empty Cup & \textbf{100} & \textbf{100} & \textbf{100} & \textbf{100} & \textbf{100} & \textbf{100} & \textbf{100} & \textbf{100} & \textbf{100} & 99 \\
        Place Fan & \textbf{99} & 95 & 96 & 95 & 96 & \textbf{96} & \textbf{99} & 93 & 87 & 85 \\
        Place Mouse Pad & \textbf{100} & \textbf{99} & 95 & 96 & 83 & 89 & 93 & 96 & 60 & 39 \\
        Place Object Basket & 88 & \textbf{92} & \textbf{93} & 91 & 89 & 88 & 91 & 88 & 80 & 76 \\
        Place Object Scale & \textbf{99} & \textbf{98} & 90 & 96 & 90 & 97 & 96 & 95 & 86 & 80 \\
        Place Object Stand & \textbf{100} & \textbf{100} & 98 & \textbf{100} & 90 & 94 & 99 & 96 & 91 & 85 \\
        Place Phone Stand & 97 & 96 & \textbf{99} & 96 & 97 & \textbf{99} & 97 & 97 & 81 & 81 \\
        Place Shoe & 99 & \textbf{100} & \textbf{100} & \textbf{100} & 96 & 99 & 98 & 98 & 92 & 93 \\
        Press Stapler & 93 & 94 & \textbf{94} & 96 & 90 & \textbf{97} & 85 & 82 & 87 & 83 \\
        Put Bottles Dustbin & 93 & \textbf{97} & 92 & 96 & \textbf{95} & 90 & 87 & 91 & 84 & 79 \\
        Put Object Cabinet & 87 & 85 & 91 & \textbf{89} & \textbf{94} & \textbf{89} & 85 & 87 & 80 & 79 \\
        Rotate QRcode & 94 & \textbf{95} & \textbf{97} & 93 & 93 & 89 & 96 & 91 & 89 & 87 \\
        Scan Object & \textbf{97} & \textbf{92} & 96 & 87 & 89 & \textbf{92} & 96 & 91 & 72 & 65 \\
        Shake Bottle & \textbf{100} & \textbf{100} & \textbf{100} & \textbf{100} & \textbf{100} & \textbf{100} & \textbf{100} & 97 & 99 & 97 \\
        Shake Bottle Horizontally & \textbf{100} & \textbf{100} & \textbf{100} & \textbf{100} & \textbf{100} & \textbf{100} & \textbf{100} & 99 & 99 & 99 \\
        Stack Blocks Three & 97 & 97 & 98 & 97 & 95 & 97 & \textbf{99} & \textbf{98} & 91 & 76 \\
        Stack Blocks Two & \textbf{100} & \textbf{100} & \textbf{100} & \textbf{100} & \textbf{100} & \textbf{100} & \textbf{100} & 98 & 97 & \textbf{100} \\
        Stack Bowls Three & 80 & 80 & \textbf{89} & \textbf{87} & 80 & 81 & 86 & 83 & 77 & 71 \\
        Stack Bowls Two & \textbf{96} & 97 & 92 & 96 & 92 & \textbf{98} & 94 & \textbf{98} & 95 & 96 \\
        Stamp Seal & \textbf{100} & \textbf{100} & \textbf{100} & 99 & 90 & 94 & 96 & 97 & 79 & 55 \\
        Turn Switch & 50 & \textbf{62} & 55 & 60 & 61 & 59 & 44 & 45 & \textbf{62} & 54 \\
        \midrule
        \textbf{Average} & \textbf{95.60} & \textbf{95.52} & 94.22 & 94.20 & 91.88 & 91.78 & 92.90 & 91.50 & 82.74 & 76.76 \\
        \bottomrule
    \end{tabular*}
\end{table}

\clearpage
\section{Real-World Experimental Details}
\label{app:real_world}

\subsection{Robot Platform and Observations}
\label{app:real_world_hardware}
\begin{figure}[!tb]
    \centering
    \includegraphics[width=\columnwidth,trim=430 270 490 330,clip]{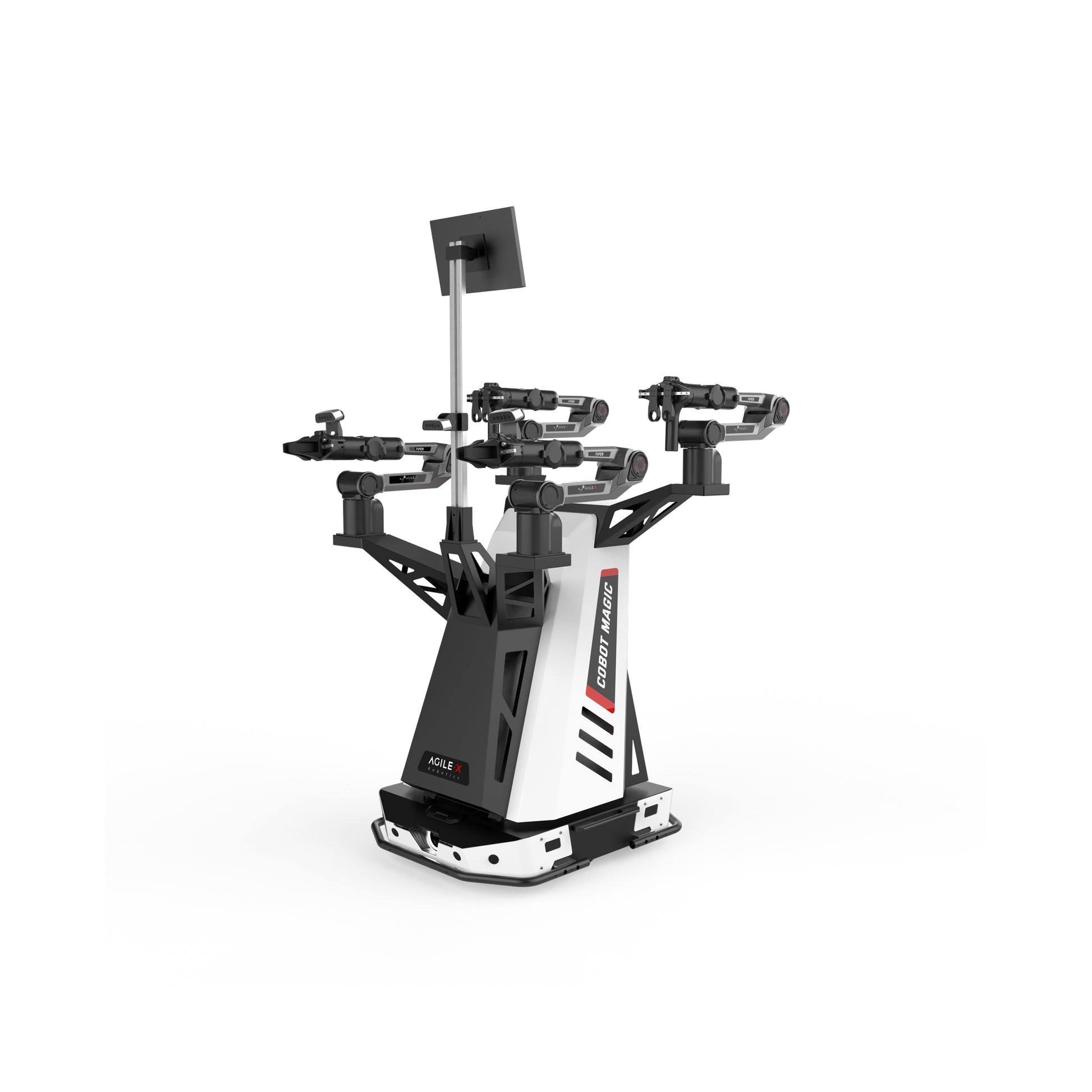}
    \caption{AgileX Cobot Magic platform. Two follower arms perform bimanual manipulation, while two leader arms support demonstration collection. The observation configuration comprises one scene camera and two wrist cameras. Hardware reference image from \href{https://global.agilex.ai/products/cobot-magic}{AgileX Robotics}.}
    \label{fig:real_world_platform}
\end{figure}

Experiments are conducted on an AgileX (Songling) Cobot Magic platform equipped with two six-degree-of-freedom PiPER arms and parallel-jaw grippers. Figure~\ref{fig:real_world_platform} illustrates the hardware platform. Manipulation takes place in a fixed tabletop workspace. Visual observations comprise three RGB views: a scene camera covering the shared workspace and a wrist-mounted camera on each arm. The scene view provides task-level context, while the wrist views resolve local interactions between the grippers, objects, and target receptacles. Robot placement and camera mounting geometry are held fixed during data collection to maintain a consistent observation configuration.

\subsection{Demonstration Collection and Data Representation}
\label{app:real_world_data}
Human demonstrations are collected through teleoperation. Across episodes, object positions and in-plane orientations are varied within the reachable workspace, subject to the geometric constraints of each task. This variation changes the required grasps and placements while preserving the task objective. Each trajectory associates multiview visual observations with bimanual robot states, action sequences, a language instruction, and an episode outcome. Subtask descriptions and boundaries supervise action learning, planning, and state estimation.

The collection protocol retains successful, partially completed, and failed executions. Failures are annotated by the stage at which they occur and their principal cause, preserving information about completed task prefixes. Recording validity is assessed separately from task success: episodes with missing visual or robot data, unusable temporal alignment, or severe occlusion of the manipulation are excluded. Collection uses no fixed success-to-failure ratio.

Trajectories are organized in LeRobot format with episode-level metadata and aligned camera streams. Recorded videos are stored at 30 frames per second. LAW uses the end-effector action representation described in the main paper: a three-dimensional position, a unit quaternion, and a scalar gripper command for each arm, giving a 16-dimensional bimanual action vector. Semantic boundaries align visual observations, actions, and subtask labels along the trajectory timeline.

\subsection{Task Definitions}
\label{app:real_world_tasks}
The evaluation comprises three tasks with complementary requirements for sequential manipulation, spatial alignment, and state-dependent interaction.

\noindent\textbf{Domino.}
The robot arranges four dominoes into an upright row and initiates their toppling. Successful execution requires stable placement and sufficient spatial alignment for the intended toppling sequence. The five stages comprise four placements and a final toppling action.

\noindent\textbf{Puzzle.}
The robot inserts three colored pieces into their corresponding recesses in a puzzle board. Each piece constitutes one semantic stage. The task requires matching and aligning each piece before insertion; success requires seating all three pieces.

\noindent\textbf{Organize.}
The robot sorts three colored blocks into the corresponding drawers of a three-level cabinet. For each block, the robot opens the target drawer, places the block inside, and closes the drawer, resulting in nine semantic stages. Success requires correct block--drawer assignments and closure of all drawers. The task combines object manipulation with repeated changes to drawer states.

\subsection{Evaluation Metrics}
\label{app:real_world_metrics}
We distinguish complete-task success from partial progress. For a task with $N$ evaluation episodes and $K$ semantic stages, let $y_i\in\{0,1\}$ indicate whether episode $i$ satisfies the complete task objective, and let $c_{i,k}\in\{0,1\}$ indicate completion of stage $k$. We report
\begin{equation}
    \mathrm{SR}=\frac{100}{N}\sum_{i=1}^{N} y_i,
    \qquad
    \mathrm{Score}=\frac{100}{NK}\sum_{i=1}^{N}\sum_{k=1}^{K}c_{i,k}.
\end{equation}
Thus, the progress score assigns equal weight to the semantic stages within each task, with $K=5$, $3$, and $9$ for Domino, Puzzle, and Organize, respectively. A partially completed execution can receive a nonzero progress score while remaining unsuccessful under the complete-task criterion. This distinction captures intermediate progress in long-horizon execution without conflating it with achievement of the final objective.

\section{Motion-Based Progress Labels}
\label{app:progress_labels}

We derive offline progress labels within each annotated subtask interval $[a,b]$ from cumulative robot-state motion. For the arms involved in the subtask, let $\mathbf p_{\ell,k}$, $\mathbf q_{\ell,k}$, and $g_{\ell,k}$ denote end-effector position, unit quaternion, and gripper state at frame $k$. The frame-to-frame motion increments are
\begin{equation}
\begin{aligned}
\delta_k^{\mathrm{pos}}&=\sum_{\ell}\|\mathbf p_{\ell,k}-\mathbf p_{\ell,k-1}\|_2,\\
\delta_k^{\mathrm{rot}}&=\sum_{\ell}2\arccos\!\left(\operatorname{clip}(|\mathbf q_{\ell,k}^{\top}\mathbf q_{\ell,k-1}|,0,1)\right),\\
\delta_k^{\mathrm{grip}}&=\sum_{\ell}|g_{\ell,k}-g_{\ell,k-1}|.
\end{aligned}
\end{equation}
Quaternion normalization and the absolute inner product make the rotation increment invariant to quaternion sign. For component $z\in\{\mathrm{pos},\mathrm{rot},\mathrm{grip}\}$, define
\begin{equation}
C_k^z=\sum_{i=a+1}^{k}\delta_i^z,\qquad
P_k^z=\operatorname{clip}\!\left(C_k^z/C_b^z,0,1\right).
\end{equation}
Each component is normalized by its own total motion within the subtask, avoiding direct addition of quantities with different units. Components are included only when their total motion is at least $0.01$ for position, $2^\circ$ for rotation, and $0.05$ for gripper state. Position uses the recorded position units (meters for end-effector trajectories), rotation is computed in radians, and gripper motion uses the recorded gripper-state scale. These thresholds exclude near-static components whose normalization would amplify negligible motion. With $\mathcal A$ denoting the included components, progress is
\begin{equation}
P_k=\frac{1}{|\mathcal A|}\sum_{z\in\mathcal A}P_k^z.
\end{equation}
The included components receive equal weights $1/|\mathcal A|$; no learned or modality-specific weights are used. If robot states are unavailable or no component passes the motion threshold, progress falls back to clipped temporal interpolation $(k-a)/\max(1,b-a)$.

For 14-dimensional joint-state trajectories, the implementation instead uses the Euclidean change of the first three joints of each involved arm and the summed absolute change of its remaining three joints, together with the gripper change. These are joint-motion proxies rather than Cartesian translation and geodesic rotation; the same component-wise normalization and numerical inclusion thresholds apply.

Completion labels are assigned to the contiguous suffix of the annotated interval with $P_k\geq0.995$, scanning backward from $b$. If no frame meets the threshold, the annotated endpoint $b$ is retained. The threshold corresponds to the final $0.5\%$ of normalized progress and provides a narrow endpoint tolerance rather than requiring exact equality to one. It is an implementation choice, not a task-independent physical success criterion. Motion progress locates the completion region within an annotated subtask; semantic meaning is supplied by the subtask annotation rather than by motion magnitude alone.

\section{Observation and Action Conventions}
\label{app:observation_action}

In the implementation, RGB inputs comprise one scene view and two wrist views, with separate historical visual-memory streams for the VLM and WAM. The WAM receives a 16-dimensional EEF state; the VLM state expert receives textual values for six joint positions and one gripper state per arm (14 dimensions), while the planner uses visual and task context. Actions are absolute 16-dimensional EEF targets ordered as left $(x,y,z,q_w,q_x,q_y,q_z,g)$ followed by right. Absolute poses use scene-world coordinates with the controller-specific end-effector frame convention; no relative-action conversion is applied. Gripper targets use the normalized environment scale, with zero closed and one open. WAM states and actions use per-dimension global z-score statistics from the configured training datasets, cached in \texttt{dataset\_stats.json}; normalization uses $(x-\mu)/(\sigma+10^{-8})$, clipped to $[-5,5]$, rather than separate statistics for each action-horizon step. The same cached statistics are loaded for evaluation. Predicted actions are inverse-normalized before execution, and quaternion targets are renormalized to unit length. VLM joint-state values are supplied in their recorded scale.

\section{Semantic Subtask Decomposition}
\label{app:subtask_decomposition}

Tables~\ref{tab:robotwin-subtask-taxonomy} and~\ref{tab:robodojo-subtask-taxonomy} expand the examples in Fig.~\ref{fig:subtask} with complete task inventories. Each subtask specifies a semantic objective and its associated trajectory interval. Grasping, transport, and release can form one placement stage, whereas handovers and explicit confirmation actions can define separate stages. Repeated motions serving the same objective may remain within one subtask; object identities and spatial relations refer to the corresponding task instance.

\begin{table}[!tb]
    \centering
    \caption{Semantic subtask sequences for 50 RoboTwin tasks.}
    \label{tab:robotwin-subtask-taxonomy}
    \renewcommand{\arraystretch}{1.0}
    \setlength{\tabcolsep}{4pt}
    \fontsize{8}{9}\selectfont
\begin{tabular}{@{}p{0.235\textwidth}p{\dimexpr0.765\textwidth-8pt\relax}@{}}
        \toprule
        \textbf{Task} & \textbf{Ordered semantic subtasks} \\
        \midrule
        Adjust Bottle & \textbf{(1)} Pick up the target bottle and hold it upright. \\
        Beat Block Hammer & \textbf{(1)} Hit the block with the hammer. \\
        Blocks Ranking RGB & \textbf{(1)} Place the red block at the left position; \textbf{(2)} Place the green block at the middle position; \textbf{(3)} Place the blue block at the right position. \\
        Blocks Ranking Size & \textbf{(1)} Place the small block at the right position; \textbf{(2)} Place the medium block at the middle position; \textbf{(3)} Place the large block at the left position. \\
        Click Alarmclock & \textbf{(1)} Press the alarm clock button. \\
        Click Bell & \textbf{(1)} Press the top of the bell. \\
        Dump Bin Bigbin & \textbf{(1)} Pour the trash bin contents into the big bin. \\
        Grab Roller & \textbf{(1)} Lift the roller. \\
        Handover Block & \textbf{(1)} Pick up the red block; \textbf{(2)} Transfer the red block to the other hand; \textbf{(3)} Place the red block on the blue pad. \\
        Handover Mic & \textbf{(1)} Pick up the microphone; \textbf{(2)} Transfer the microphone to the other hand. \\
        Hanging Mug & \textbf{(1)} Place the mug at the center of the table; \textbf{(2)} Place the mug on the rack. \\
        Lift Pot & \textbf{(1)} Lift the pot off the table. \\
        Move Can Pot & \textbf{(1)} Place the can next to the pot. \\
        Move Pillbottle Pad & \textbf{(1)} Place the pill bottle on the pad. \\
        Move Playingcard Away & \textbf{(1)} Move the playing-card box outward. \\
        Move Stapler Pad & \textbf{(1)} Place the stapler on the target mat. \\
        Open Laptop & \textbf{(1)} Open the laptop lid. \\
        Open Microwave & \textbf{(1)} Open the microwave door to more than 70 degrees. \\
        Pick Diverse Bottles & \textbf{(1)} Pick up the two target bottles. \\
        Pick Dual Bottles & \textbf{(1)} Pick up the two target bottles. \\
        Place A2B Left & \textbf{(1)} Place the target object to the left of the reference object. \\
        Place A2B Right & \textbf{(1)} Place the target object to the right of the reference object. \\
        Place Bread Basket & \textbf{(1)} Place the bread in the breadbasket. \\
        Place Bread Skillet & \textbf{(1)} Place the bread in the skillet. \\
        Place Burger Fries & \textbf{(1)} Place the hamburger on the tray; \textbf{(2)} Place the french fries on the tray. \\
        Place Can Basket & \textbf{(1)} Place the can in the basket; \textbf{(2)} Lift the basket 5 cm above its initial height. \\
        Place Cans Plasticbox & \textbf{(1)} Place the left can in the plastic box; \textbf{(2)} Place the right can in the plastic box. \\
        Place Container Plate & \textbf{(1)} Place the container on the plate. \\
        Place Dual Shoes & \textbf{(1)} Place the left shoe in the shoe box; \textbf{(2)} Place the right shoe in the shoe box. \\
        Place Empty Cup & \textbf{(1)} Place the empty cup on the coaster. \\
        Place Fan & \textbf{(1)} Place the fan on the target mat facing the robot. \\
        Place Mouse Pad & \textbf{(1)} Place the mouse on the target mat. \\
        Place Object Basket & \textbf{(1)} Place the object in the basket; \textbf{(2)} Lift the basket. \\
        Place Object Scale & \textbf{(1)} Place the target object on the electronic scale. \\
        Place Object Stand & \textbf{(1)} Place the target object on the display stand. \\
        Place Phone Stand & \textbf{(1)} Place the phone on the phone stand. \\
        Place Shoe & \textbf{(1)} Place the shoe on the mat. \\
        Press Stapler & \textbf{(1)} Press the stapler. \\
        Put Bottles Dustbin & \textbf{(1)} Put one bottle in the dustbin; \textbf{(2)} Put another bottle in the dustbin; \textbf{(3)} Put the remaining bottle in the dustbin. \\
        Put Object Cabinet & \textbf{(1)} Grasp and lift the object; \textbf{(2)} Pull the cabinet drawer open; \textbf{(3)} Place the object near the center inside the cabinet drawer and fully release it. \\
        Rotate QRcode & \textbf{(1)} Pick up the payment sign; \textbf{(2)} Rotate the payment sign until the QR code faces the robot. \\
        Scan Object & \textbf{(1)} Pick up the scanner and the object; \textbf{(2)} Scan the object with the scanner. \\
        Shake Bottle & \textbf{(1)} Pick up the bottle; \textbf{(2)} Shake the bottle. \\
        Shake Bottle Horizontally & \textbf{(1)} Pick up the bottle; \textbf{(2)} Shake the bottle horizontally. \\
        Stack Blocks Three & \textbf{(1)} Place the red block at the center of the table as the base; \textbf{(2)} Stack the green block on the red block; \textbf{(3)} Stack the blue block on the green block. \\
        Stack Blocks Two & \textbf{(1)} Place the red block at the center of the table as the base; \textbf{(2)} Stack the green block on the red block. \\
        Stack Bowls Three & \textbf{(1)} Place one bowl as the base; \textbf{(2)} Stack another bowl inside the base bowl; \textbf{(3)} Stack the remaining bowl inside the stacked bowls. \\
        Stack Bowls Two & \textbf{(1)} Place one bowl as the base; \textbf{(2)} Stack the remaining bowl inside the base bowl. \\
        Stamp Seal & \textbf{(1)} Stamp the target area with the seal. \\
        Turn Switch & \textbf{(1)} Activate the switch. \\
        \bottomrule
    \end{tabular}
\end{table}

\clearpage
\begin{table}[!tb]
    \centering
    \caption{Semantic subtask sequences for 35 RoboDojo tasks.}
    \label{tab:robodojo-subtask-taxonomy}
    \renewcommand{\arraystretch}{1.0}
    \setlength{\tabcolsep}{4pt}
    \fontsize{8}{9}\selectfont
\begin{tabular}{@{}p{0.235\textwidth}p{\dimexpr0.765\textwidth-8pt\relax}@{}}
        \toprule
        \textbf{Task} & \textbf{Ordered semantic subtasks} \\
        \midrule
        Arrange Largest Number & \textbf{(1)} Arrange the digits from left to right to form the largest number. \\
        Build Tower & \textbf{(1)} Place the two base blocks of the first tower layer; \textbf{(2)} Place the first-layer beam across the base blocks; \textbf{(3)} Place the two base blocks of the second tower layer; \textbf{(4)} Place the second-layer beam across the second-layer base; \textbf{(5)} Place the cuboid block on the tower; \textbf{(6)} Place the pyramid on top to complete the tower. \\
        Classify Objects & \textbf{(1)} Sort objects into the white basket; \textbf{(2)} Sort objects into the blue basket; \textbf{(3)} Sort objects into the red basket. \\
        Cover Blocks & \textbf{(1)} Cover the leftmost block; \textbf{(2)} Cover the middle block; \textbf{(3)} Cover the rightmost block; \textbf{(4)} Uncover the red block; \textbf{(5)} Uncover the green block; \textbf{(6)} Uncover the blue block. \\
        Deposit Coin & \textbf{(1)} Pick up the coin; \textbf{(2)} Hand the coin to the other hand; \textbf{(3)} Insert the coin into the coin bank. \\
        DLC & \textbf{(1)} Arrange the letters to spell ``RoboDojo'' in a row. \\
        Fasten Screws & \textbf{(1)} Insert the first screw and tighten it with five turns; \textbf{(2)} Insert the second screw and tighten it with five turns; \textbf{(3)} Insert the third screw and tighten it with five turns. \\
        Fill Egg Holder & \textbf{(1)} Place all eggs into the egg holder; \textbf{(2)} Close the egg-holder lid. \\
        Fill Pen Holder & \textbf{(1)} Pick up the pen holder; \textbf{(2)} Place all pens into the pen holder; \textbf{(3)} Put the pen holder back down on the table. \\
        Fold Clothes & \textbf{(1)} Fold the left sleeve; \textbf{(2)} Fold the right sleeve; \textbf{(3)} Fold the hem. \\
        Hang Mugs & \textbf{(1)} Hang all mugs on the mug rack. \\
        Imitate Sorting Sequence & \textbf{(1)} Observe and remember the demonstrated placement sequence; \textbf{(2)} Place all objects following the demonstrated sequence. \\
        Insert Key & \textbf{(1)} Pick up the key; \textbf{(2)} Hand the key to the other hand; \textbf{(3)} Insert the key into the keyhole and turn it. \\
        Insert Tubes & \textbf{(1)} Insert all tubes into the rack. \\
        Make Kong & \textbf{(1)} Wait for the opponent to discard a matching tile; \textbf{(2)} Declare a kong with the matching tiles; \textbf{(3)} Draw a replacement tile. \\
        Make Toast & \textbf{(1)} Place both bread slices into the toaster; \textbf{(2)} Press the toaster lever down. \\
        Match And Pick From Conveyor & \textbf{(1)} Observe and remember the first object on the conveyor; \textbf{(2)} Pick the matching object when it reappears. \\
        Organize Table & \textbf{(1)} Place the alarm clock on the drawer; \textbf{(2)} Place the figurine on the stand; \textbf{(3)} Place the mouse on the mouse pad; \textbf{(4)} Push the keyboard into the frame. \\
        Pack Objects Into Box & \textbf{(1)} Pack the objects into the box with their front sides facing left. \\
        Play Xylophone & \textbf{(1)} Pick up the mallet; \textbf{(2)} Strike all xylophone keys from left to right. \\
        Play Stacking Toy & \textbf{(1)} Place all pieces of the first stacking toy onto their peg; \textbf{(2)} Place all pieces of the second stacking toy onto their peg; \textbf{(3)} Place all pieces of the third stacking toy onto their peg; \textbf{(4)} Place all pieces of the fourth stacking toy onto their peg. \\
        Play Tic Tac Toe & \textbf{(1)} Place all robot marks on the tic-tac-toe board. \\
        Plug In Charger & \textbf{(1)} Pick up the charger; \textbf{(2)} Hand the charger to the other hand; \textbf{(3)} Plug the charger into the power strip. \\
        Pour Balls Into Vase & \textbf{(1)} Pour all balls from the cup into the vase. \\
        Pour Liquid Into Cup & \textbf{(1)} Pour the liquid from the bottle into the cup. \\
        Press By Number & \textbf{(1)} Press the first red button the required number of times; \textbf{(2)} Press the blue button to confirm; \textbf{(3)} Press the second red button the required number of times; \textbf{(4)} Press the blue button to confirm. \\
        Push T & \textbf{(1)} Push the T-shaped block into alignment with the gray T-shaped pad. \\
        Put Bottles Into Dustbin & \textbf{(1)} Put all bottles into the dustbin. \\
        Sort Nesting Dolls By Size & \textbf{(1)} Place all nesting dolls in size order from left to right, smallest to largest. \\
        Stack Blocks & \textbf{(1)} Place the first block as the tower base; \textbf{(2)} Stack the second block on the base; \textbf{(3)} Stack the final block to complete the tower. \\
        Stack Bowls & \textbf{(1)} Stack the three bowls together. \\
        Store Laptop And Headphones & \textbf{(1)} Hang the headphones on the headphone stand; \textbf{(2)} Close the laptop; \textbf{(3)} Place the laptop into the vertical laptop stand. \\
        Swap T & \textbf{(1)} Pick up the first T-shaped block; \textbf{(2)} Pick up the second T-shaped block; \textbf{(3)} Place the first T-shaped block at the second T's original position; \textbf{(4)} Place the second T-shaped block at the first T's original position. \\
        Swap Blocks & \textbf{(1)} Move the first block to the empty mat; \textbf{(2)} Press the button after moving the first block; \textbf{(3)} Move the second block to the first block's original position; \textbf{(4)} Press the button after moving the second block; \textbf{(5)} Move the first block to the second block's original position; \textbf{(6)} Press the button to complete the swap. \\
        Sweep Blocks & \textbf{(1)} Pick up the broom; \textbf{(2)} Hand the broom to the right hand; \textbf{(3)} Sweep the blocks into the dustpan. \\
        \bottomrule
    \end{tabular}
\end{table}

\subsection{Real-World Subtask Sequences}
Table~\ref{tab:real-world-subtask-sequences} specifies the five, three, and nine stages for Domino, Puzzle, and Organize, respectively, consistent with the task definitions and progress metric in Appendix~\ref{app:real_world}.

\begin{table}[!tb]
    \centering
    \caption{Semantic subtask sequences for real-world tasks.}
    \label{tab:real-world-subtask-sequences}
    \renewcommand{\arraystretch}{1.0}
    \setlength{\tabcolsep}{4pt}
    \fontsize{8}{9.2}\selectfont
\begin{tabular}{@{}p{0.235\textwidth}p{\dimexpr0.765\textwidth-8pt\relax}@{}}
        \toprule
        \textbf{Task} & \textbf{Ordered semantic subtasks} \\
        \midrule
        Domino & \textbf{(1--4)} Place the first, second, third, and fourth dominoes upright in sequence; \textbf{(5)} initiate the toppling sequence. \\
        Puzzle & \textbf{(1)} Insert the first piece into its matching recess; \textbf{(2)} insert the second piece; \textbf{(3)} insert the third piece. \\
        Organize & \textbf{(1)} Open the drawer for block 1; \textbf{(2)} place block 1 inside; \textbf{(3)} close the drawer; \textbf{(4)} open the drawer for block 2; \textbf{(5)} place block 2 inside; \textbf{(6)} close the drawer; \textbf{(7)} open the drawer for block 3; \textbf{(8)} place block 3 inside; \textbf{(9)} close the drawer. \\
        \bottomrule
    \end{tabular}
\end{table}

\clearpage
\subsection{Segmented Trajectory Examples}
\begin{figure}[!tb]
    \centering
    \includegraphics[width=\textwidth]{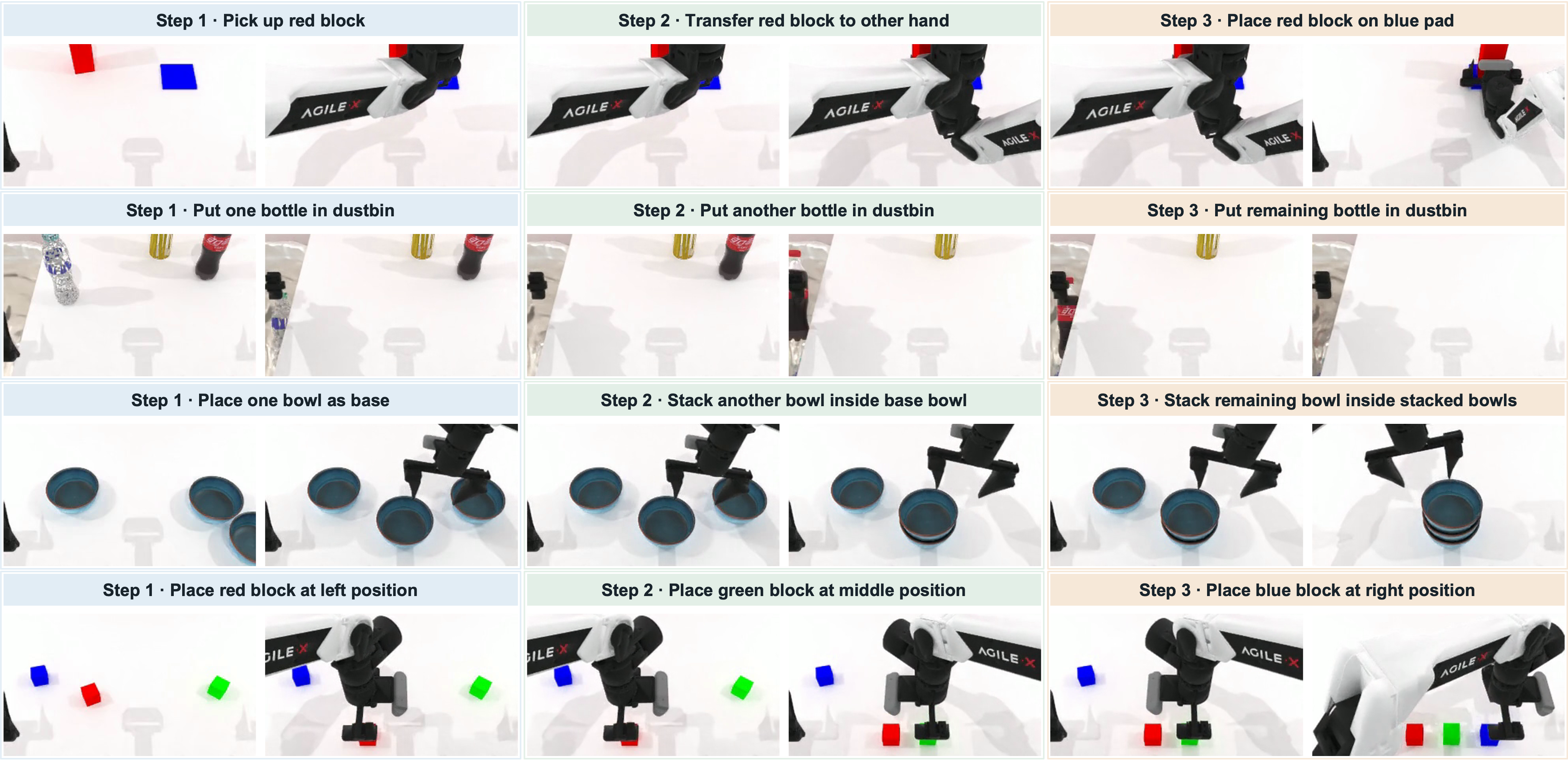}
    \caption{Semantic subtask segmentation on RoboTwin. From top to bottom: Handover Block, Put Bottles Dustbin, Stack Bowls Three, and Blocks Ranking RGB. The examples distinguish acquisition, inter-arm transfer, and placement, as well as successive placements involving different objects.}
    \label{fig:subtask-robotwin-1}
\end{figure}

\begin{figure}[!tb]
    \centering
    \includegraphics[width=\textwidth]{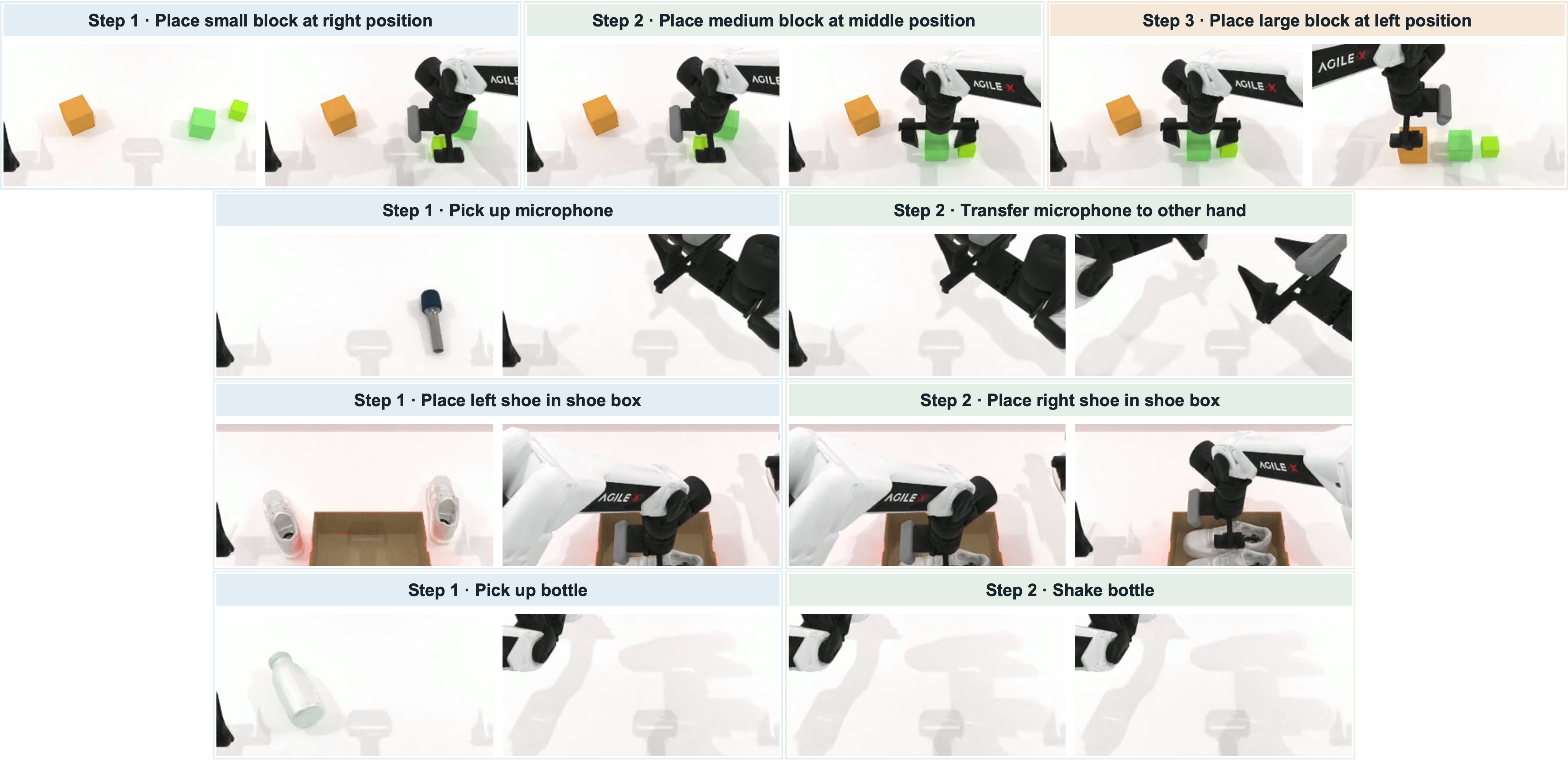}
    \caption{Additional RoboTwin examples. From top to bottom: Blocks Ranking Size, Handover Mic, Place Dual Shoes, and Shake Bottle. The stage boundaries separate size-dependent placements, an inter-arm transfer, left- and right-shoe placements, and bottle acquisition from shaking.}
    \label{fig:subtask-robotwin-2}
\end{figure}

\begin{figure}[!tb]
    \centering
    \includegraphics[width=\textwidth]{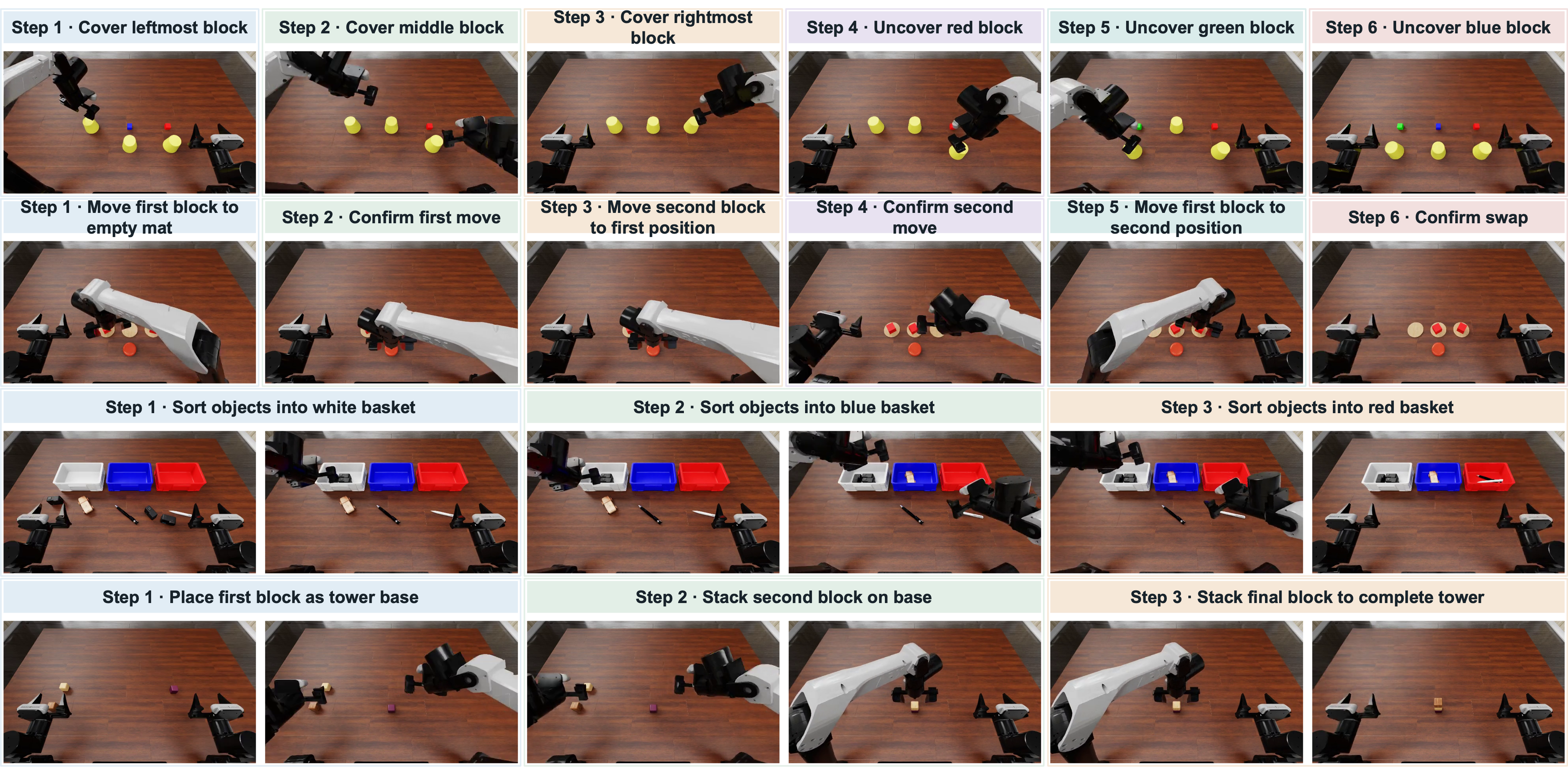}
    \caption{Semantic subtask segmentation on RoboDojo. From top to bottom: Cover Blocks, Swap Blocks, Classify Objects, and Stack Blocks. Covering and uncovering are represented as distinct stages; in Swap Blocks, each confirmation stage corresponds to a button press following a transfer. Sorting is grouped by the destination basket.}
    \label{fig:subtask-robodojo-1}
\end{figure}

\begin{figure}[!tb]
    \centering
    \includegraphics[width=\textwidth]{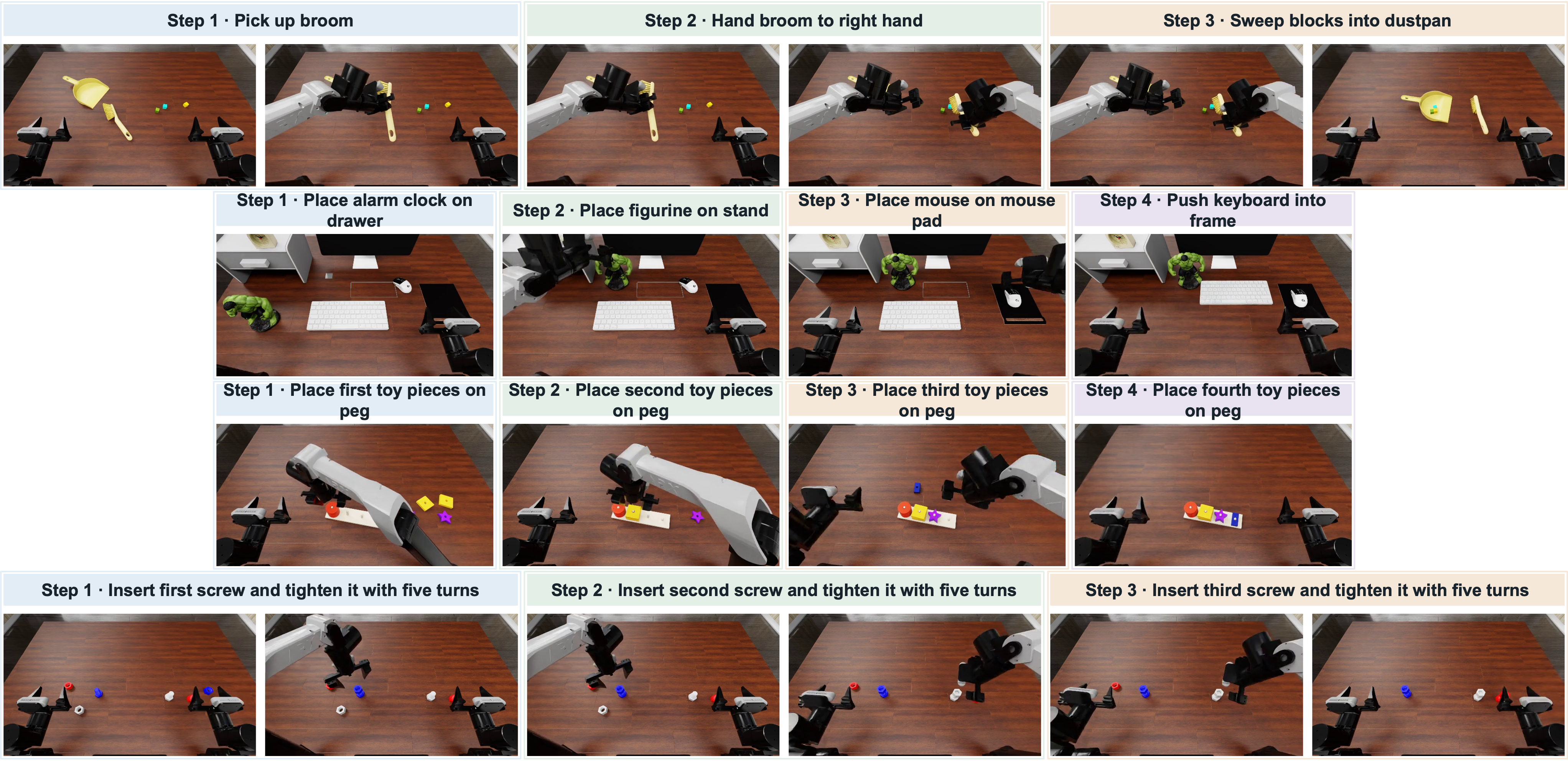}
    \caption{Additional RoboDojo examples. From top to bottom: Sweep Blocks, Organize Table, Play Stacking Toy, and Fasten Screws. The decompositions preserve tool acquisition and transfer, object-specific placement objectives, and repeated fastening operations. In Fasten Screws, insertion and five tightening turns form one subtask per screw.}
    \label{fig:subtask-robodojo-2}
\end{figure}

Figures~\ref{fig:subtask-robotwin-1}--\ref{fig:subtask-robodojo-2} provide additional examples from the simulation datasets. Each row corresponds to a task, and the labeled groups delineate semantic subtasks in execution order from left to right. One or two representative observations depict each stage. The grouping follows changes in the semantic objective rather than uniform temporal sampling.


\clearpage

\end{document}